\documentclass[11pt]{article}
\pdfoutput=1
\usepackage{latexsym}
\usepackage[tbtags]{amsmath}
\usepackage{amssymb}
\usepackage{amsmath}
\usepackage{color}
\usepackage{graphicx}
\usepackage{authblk}
\usepackage{cite}
\usepackage[usenames,dvipsnames]{xcolor}
\usepackage[pdfborder={0 0 0}]{hyperref}
\usepackage{algorithm} 
\usepackage{algorithmic}
\usepackage{lipsum}
\usepackage{tablefootnote}
\usepackage{fixltx2e}

\allowdisplaybreaks
\def\baselinestretch{1.5}

\def\inclde-picture #1 by #2 (#3){%
	\vbox to #2{%
		\hrule width #1 height 0pt depth 0pt
		\vfill
		\special{picture #3}}}
\def\scaledpicture #1 by #2 (#3 scaled #4){{%
		\dimen0=#1 \dimen1=#2
		\divide\dimen0 by 1000 \multiply \dimen0 by #4
		\divide\dimen1 by 1000 \multiply \dimen1 by #4
		\inclde-picture \dimen0 by \dimen1 (#3 scaled #4)}}

\def\inclde-picture #1 by #2 (#3){%
	\vbox to #2{%
		\hrule width #1 height 0pt depth 0pt
		\vfill
		\special{picture #3}}}
\def\scaledpicture #1 by #2 (#3 scaled #4){{%
		\dimen0=#1 \dimen1=#2
		\divide\dimen0 by 1000 \multiply \dimen0 by #4
		\divide\dimen1 by 1000 \multiply \dimen1 by #4
		\inclde-picture \dimen0 by \dimen1 (#3 scaled #4)}}
\newcounter{example}

\newcounter{para}
  
\newtheorem{definition}{Definition}[section]

\newcounter{remarkcnt}
\newcommand{\Lower}[1]{\smash{\lower 1.5ex \hbox{#1}}}
\newcommand{\HLower}[1]{\smash{\lower .03in \hbox{#1}}}
\newcommand{\LHigher}[1]{\smash{\raise .02in \hbox{#1}}}

\usepackage{natbib}
\usepackage[caption=false]{subfig}
\usepackage{tablefootnote}
\usepackage{threeparttable}
\usepackage{multirow}

\usepackage{hyperref}
\hypersetup{
	colorlinks=true,
	linkcolor=blue,
	filecolor=magenta,      
	urlcolor=cyan,
}
\usepackage{booktabs}
\usepackage{graphicx}
\usepackage{xr}
\usepackage{filecontents}
\usepackage{lscape}
\usepackage{mathtools}
\usepackage{ulem}

\begin{document}
	\makeatletter\@input{zzAppendicesAUXfile.tex}\makeatother
	\setcounter{footnote}{1}
	
	\vspace{.6in}
	\title{\LARGE Learning Optimal Solutions via an LSTM-Optimization Framework}

	\author[1]{Dogacan Yilmaz}
	\author[2]{\.I. Esra B\"uy\"uktahtak{\i}n\thanks{Corresponding author email: esratoy@vt.edu}}
	\affil[1]{Department of Mechanical and Industrial Engineering, New Jersey Institute of Technology}
	\affil[2]{Department of Industrial and Systems Engineering, Virginia Tech}
	
	\maketitle
	
	\renewcommand\baselinestretch{1.5}

	\abstract
	
	In this study, we present a deep learning-optimization framework to tackle dynamic mixed-integer programs. Specifically, we develop a bidirectional Long Short Term Memory (LSTM) framework that can process information forward and backward in time to learn optimal solutions to sequential decision-making problems. We demonstrate our approach in predicting the optimal decisions for the single-item capacitated lot-sizing problem (CLSP), where a binary variable denotes whether to produce in a period or not. Due to the dynamic nature of the problem, the CLSP can be treated as a sequence labeling task where a recurrent neural network can capture the problem's temporal dynamics. Computational results show that our LSTM-Optimization (LSTM-Opt) framework significantly reduces the solution time of benchmark CLSP problems without much loss in feasibility and optimality. For example, the predictions at the 85\% level reduce the CPLEX solution time by a factor of 9 on average for over 240,000 test instances with an optimality gap of less than 0.05\% and 0.4\% infeasibility in the test set. Also, models trained using shorter planning horizons can successfully predict the optimal solution of the instances with longer planning horizons.  For the hardest data set, the LSTM predictions at the 25\% level reduce the solution time of 70 CPU hours to less than 2 CPU minutes with an optimality gap of 0.8\% and without any infeasibility. The LSTM-Opt framework outperforms classical ML algorithms, such as the logistic regression and random forest, in terms of the solution quality, and exact approaches, such as the ($\ell$, S) and dynamic programming-based inequalities, with respect to the solution time improvement. Our machine learning approach could be beneficial in tackling sequential decision-making problems similar to CLSP, which need to be solved repetitively, frequently, and in a fast manner.

	\noindent Keywords: Machine learning, recurrent neural networks, bidirectional Long Short Term Memory (LSTM), mixed integer programming, capacitated lot-sizing, sequential decision-making


\section{Introduction}

In the recent decade, significant progress has been achieved with the use of machine learning (ML) in various fields, such as image recognition and natural language processing. A subfield of ML, deep learning, has inspired much success over the last decade and has led to a growing interest in research and practice. ML and operations research (OR) are historically interconnected through optimization, but only recently the use of ML for OR has received more attention. In this study, we will focus on this direction. Specifically, we will leverage deep learning algorithms to predict solutions to an OR problem by taking advantage of previously solved problems. In various applications in operations planning and management, such as energy demand-side management, airline scheduling, and vehicle routing, problems with the same structures must be solved repeatedly with different parameters within a very short period of time. In such settings, a reduced solution time obtained by fast algorithms can be highly beneficial for improving the efficiency and performance of businesses.

One complex and recurring problem for industrial companies is to determine the amount and timing of production over a planning horizon under resource constraints. It is an important challenge in industry and supply chain management because a production plan directly impacts companies’ output and their ability to compete in operational costs and customer service levels \citep{gicquel2008capacitated}. Production planning is also a highly complex task because firms strive to optimize multiple conflicting objectives, such as minimizing production and inventory costs, while maximizing customer satisfaction under tight constraints on resources, such as budget, raw materials, and machine availability. 

In this paper, we present a general prediction framework to learn optimal solutions of combinatorial optimization problems, while focusing on tackling one core production planning problem: the single-item Capacitated Lot Sizing Problem (CLSP). The practical importance of the CLSP is apparent from numerous examples of its application in various production and manufacturing industries, including but not limited to the textile industry, oil and gas companies, car manufacturers, and pharmaceutical industry \citep{karimi2003capacitated,gicquel2008capacitated}. The CLSP determines the optimal production and inventory levels that meet periodic demand under a given production capacity by minimizing the sum of production, setup, and inventory holding cost over a finite planning horizon. In the mixed-integer programming (MIP) formulation of the CLSP, the decision of whether to produce or not is represented by a binary variable. Thus, the CLSP with time-varying capacity is NP-hard, a very difficult problem to optimize \citep{bitran1982computational,hartman2010dynamic}. In this paper, we focus on tackling the computational difficulty of the CLSP and provide an ML-based optimization framework to solve its MIP formulation more efficiently. Thus, we study the CLSP at a high formulation level rather than focusing on a specific real-life application.

The CLSP is a sequential decision-making problem because, in each time period, the production level is determined to meet the periodic demand, and any additional produced items not used for current demand are placed in inventory to be used for future demand. Therefore, demand, capacity, cost inputs, and production decisions constitute highly correlated temporal sequences that the classical supervised classification might not capture. Thus, the CLSP can be treated as a sequence labeling task where a recurrent neural network (RNN) is applicable.

The RNN is a specialized type of neural network that can process sequential data by enabling information flow through various time steps. The neural network with the same parameters is applied at each time step of the sequence. Input to layer at each time step consists of data at that time step and the network activations from the previous step. As a result, RNNs allow previous inputs to affect the output rather than just the current input. Developed by \citet{sepp1997long}, Long Short-Term Memory (LSTM) is a specialized RNN that can store information for long time steps, which can be challenging to handle by a classical RNN \citep{bengio1994learning}. Bidirectional RNN (BRNN) allows using input information of future time steps rather than processing information in sequential order \citep{schuster1997bidirectional}. The main idea is to train two separate RNNs in both time directions that connect to the same output layer. Bidirectional LSTM is an extension of BRNNs by using LSTM architecture \citep{graves2015framewise}. The LSTM architecture might be preferable to the classical RNN due to its ability to capture long-term dependencies that come at a computational cost. We train the bidirectional LSTM network on datasets with different characteristics and evaluate the quality of resulting predictions in terms of feasibility and optimality. Our computational results show that a significant reduction in solution time can be achieved without much loss in feasibility and optimality.

Our main contribution is to develop an LSTM-based framework for learning optimal solutions to CLSP quickly. We propose using bidirectional LSTM to predict the binary production decision variable. Bidirectional LSTM can process information in both time directions, which is critical to predicting solutions of various OR problems with dynamic nature and where data is available for the planning horizon. Also, instead of using all the predicted variables, we propose using them partially to reduce the number of infeasible solutions. We present the results on how the quality of the predictions changes regarding feasibility and optimality with different levels of predicted variables. Additionally, we show the results of generalization on instances with different characteristics and instances with longer time horizons and compare them with those of traditional dynamic programming and cutting plane algorithms and well-known learning algorithms: logistic regression and random forest.

\section{Literature Review and Contributions}

\subsection{Literature Review}

In recent years, significant results were achieved in various fields by deep learning, which is a sub-field of ML. As a result, there has been a growing literature on the interaction between ML and OR. In this paper, we focus on the use of ML to improve solving OR problems, particularly focusing on the CLSP. The origin of the interest in using ML algorithms for OR can be traced back to the 1980s when Neural Computation was used for solving Combinatorial Optimization problems (see, e.g., the survey of \citet{smith1999neural} on this topic). The approaches in the literature are structured into two parts: approaches that use ML for predicting the solutions directly from inputs and approaches that predict valuable pieces of information to utilize in the solution algorithms.

In one of the studies, which focuses on predicting the optimal solution directly, \citet{larsen2021predicting} propose a new methodology to predict solution descriptions of a stochastic load planning problem using deep learning. According to the authors, a solution can be described at different levels. The most detailed solution describes the values taken by each variable, and the least detailed solution gives the value of the objective function. Their desired level of description is somewhere in the middle. At the time of the prediction, using a deterministic optimization model is not possible because the information available is imperfect, and the computational budget is limited. They generate training data by solving a large number of deterministic problems offline and combining solutions to the desired level of description at the prediction time. They train feedforward neural networks using this generated data and predict the actual problem instances. With a similar approach, \citet{fischetti2019machine} use various ML techniques, including neural networks, to estimate the optimal value of the offshore wind farm layout optimization problem. Their goal is to determine the optimal allocation of the wind turbines in a site to maximize park power production. The authors argue that ML can be used as a fast tool to estimate the optimal value of the problem for pre-selecting between candidate wind farm sites. The optimization model can be evaluated at these promising sites instead of all candidates. Based on their findings, a fast ML+OR tool can dramatically increase the number of sites and turbine types investigated. 

In a recent study, \citet{bertsimas2019online} use neural networks to exploit the repetitive nature of online optimization, where problems with different parameters are solved frequently. They utilize the structure of mixed-integer quadratic optimization problems using neural networks to predict the strategy, which is defined as a tuple of indexes of tight inequality constraints and values of the integer variables. At the time of prediction, they do not require a solver. They evaluate a single neural network prediction and a single linear system solution.

\citet{oroojlooyjadid2019applying} utilize deep neural networks to determine the optimal order quantity in the newsvendor problem. They establish an algorithm that integrates demand forecasting with deciding optimal order quantity rather than doing both separately. The input data consists of features of demand, and the output is the optimal order quantities. Additionally, they modify the loss function of the neural network as the newsvendor objective. 


In the group of studies where ML is used to generate vital information to use in solution algorithms, \citet{khalil2016learning} propose an ML framework for strong branching decisions, leading to significantly smaller search trees. In \citet{khalil2017learning}, authors use ML to decide if a primal heuristic should be run at which nodes during the branch-and-bound tree search so that the overall performance of the solver is optimized. The reader is referred to the survey of \citet{lodi2017on} on learning algorithms to improve branch-and-bound decisions. \citet{xavier2019learning} propose the usage of ML algorithms to improve the computational performance of MIP solvers by predicting redundant constraints, reasonable initial feasible solutions, and affine subspaces where the optimal solution is likely to lie. \citet{kruber2017learning} address whether or not a reformulation should be performed and which decomposition method to choose when several are possible using ML algorithms. \citet{bonami2018learning} suggest a methodology that determines the linearization decision for a mixed-integer quadratic programming problem.

The CLSP has been widely studied in the OR literature by developing exact and heuristic algorithms. \citet{florian1980deterministic} provide a solution methodology based on dynamic programming (DP) for lot-sizing. An exact solution approach presented by \citet{barany1984strong} involves generating valid ($\ell$,S) inequalities and adding them to the formulation with a separation algorithm. \citet{eppen1987solving} redefine variables to generate a graph representation of the problem which has a tighter linear relaxation than the original formulation. More recently, DP-based and partial-objective inequalities have been proposed for the single-item CLSP \citep{hartman2010dynamic} and multi-item CLSP \citep{buyuktahtakin2018partial}, respectively. For a detailed discussion of the exact and heuristic approaches to different versions of the lot-sizing problem, we refer the readers to the excellent review of \cite{pochet2006production}.

Readers are referred to \citet{Goodfellow-et-al-2016} for a detailed discussion on deep learning algorithms. We refer to \citet{graves2012} for a detailed discussion on RNN, LSTM, and sequence labeling. Readers are referred to \citet{karimi2003capacitated} and \citet{pochet2006production} for an extensive survey on the capacitated lot-sizing problem, their variants, and exact and heuristic approaches for their solution.

There has been a growing interest in using ML algorithms to help solve OR problems in recent years. Despite all the advancements in the ML-OR integration, there is still a research gap in learning optimal solutions to MIP problems, including CLSP from previously-solved instances and evaluating the effectiveness and generalization of the learning-based optimization approach.

\subsection{Key Contributions of the Paper}

To our knowledge, none of the former studies have used a deep learning algorithm, such as LSTM, to capture the sequential nature of multi-period MIP models and predict their optimal solution. Decisions are closely linked over multiple periods in a multi-stage or sequential problem. Thus an ML approach that does not consider patterns across time may not capture the dynamic nature of the problem. The LSTM, on the other hand, is a recurrent network capable of understanding long and short-term dependencies and temporal differences in the data of optimal solutions given specific problem characteristics.

In this study, we present a new deep learning LSTM-Optimization (LSTM-Opt) architecture to learn the optimal solutions for one of the most famous combinatorial optimization problems and a classical example of a sequential decision-making problem, CLSP. Our goal here is to reduce the solution time, where numerous similar CLSP need to be solved repetitively and in a fast manner with a small optimality gap. Our specific contributions are:

\begin{enumerate}
\item To our knowledge, this is the first study that utilizes an LSTM approach to make predictions from the optimal solutions of CLSP instances and use those predictions to solve similar CLSP with different data. Specifically, we propose an LSTM-Opt framework, which predicts binary decision variables of the CLSP problem. The bidirectional LSTM learns optimal solutions to sequential decision-making problems where the input data is available for the planning horizon. We compare the computational performance of our algorithm with other ML approaches, such as logistic regression and random forest. We show that the LSTM networks capture the time-wise dependency in sequential decision making and thus are superior compared to those ML algorithms.
\item We evaluate the effectiveness of predictions in terms of their feasibility and optimality for the original CLSP by defining optimization-based metrics, such as the optimality gap and the percent of feasibly-predicted instances in the test set. The use of all predictions could help reduce the solution time but also may increase the infeasibility in the test set. To improve the feasibility of the solutions, we propose using the predictions partially as an input into the MIP solver, CPLEX \citep{ilogcplex}. This approach provides a significant reduction in solution time while improving the optimality gap and the feasibility of solutions. To remedy the infeasibility problem, additional methods, such as the CPLEX user cuts, are utilized to solve the problem with a reasonable optimality gap with no infeasibility.
\item We utilize benchmark CLSP instances in the literature to demonstrate the efficiency of our LSTM-Opt approach. In addition to comparing with direct solutions of CLSP by CPLEX, we utilize a dynamic programming formulation \citep{florian1980deterministic}, dynamic programming-based inequalities \citep{hartman2010dynamic}, and ($\ell$,S) inequalities \citep{barany1984strong} to show that the LSTM-Opt can be beneficial to reduce the solution time even when compared with these traditional exact OR methodologies proposed for solving the CLSP more efficiently.
\item Our LSTM-Opt framework helps decrease the CPLEX solution time by multiple orders of magnitude when predicting CLSP instances. Furthermore, this prediction architecture provides more time-gain benefits as the CLSP instances get harder, i.e., for the most difficult test problems that are generated with the same distribution as training instances, the solution time is reduced by a factor of 13 without any infeasibility or an optimality gap. 
\item We investigate if the trained LSTM model can predict instances with different underlying data distributions or instances with a larger planning horizon. The results imply that one must be careful in picking the prediction level to solve instances with different characteristics. The computational results also show that the trained LSTM model can successfully predict longer and thus harder instances without extra training. As an example, in those generalization experiments to predict longer planning horizons, using a prediction level of 25\%, we have reduced an average solution time of 70 CPU hours to only 2 CPU minutes with a 0.8\% optimality gap, which is a quite significant computational achievement.
\item Once an LSTM model is trained from previously solved instances, predictions to new problems can be generated in milliseconds in an online setting. Thus, our LSTM-Opt approach could, in particular, be useful for solving practical and recurring sequential decision-making problems, such as power generation scheduling, energy demand-side management, and pricing optimization, where the same problem formulations are solved repeatedly over time with updated parameters.
\item Our LSTM-Opt framework is generalizable since it does not assume any specific information about CLSP. Thus it can be applied to other MIPs, such as the Binary Knapsack problem, one of the most well-known MIP formulations and a relaxation of the CLSP.
\end{enumerate}

The remainder of the paper is as follows. Section \ref{Capacitated Lot Sizing Problem} presents the MIP formulation of the CLSP. Section \ref{LSTM Framework} describes the proposed LSTM-Opt framework. Section \ref{Implementation and Experimentation}	describes the details of implementation and experimentation. Section \ref{Results} presents the computational results on datasets with different characteristics and a comparison with other ML and exact approaches. Section \ref{Conclusions and Future Work} concludes the paper with future research directions. Appendix \ref{Results for Training LSTM Models}-\ref{Predicting Instances with Different Distributions} provides a discussion on the LSTM training time and more results with different datasets and characteristics, respectively.

\section{Capacitated Lot Sizing Problem}
\label{Capacitated Lot Sizing Problem}

CLSP is a fundamental problem in production planning. The CLSP determines the production and inventory levels in a multi-period planning horizon to fulfill the deterministic demand without back-ordering to minimize the sum of production, setup, and inventory holding costs. The CLSP with time-varying capacity is NP-Hard, and it has numerous variations and applications in the production and manufacturing industries \citep{quadt2008capacitated}.

To formulate the CLSP as a mixed-integer program (MIP), the following parameters and decision variables are defined. Let $T$ be the number of periods considered in the planning horizon. For each period $t \in \left\{1,2,\ldots,T \right\}$ demand $d_{t}$ is known in advance. For each period $t \in \left\{1,2,\ldots,T \right\}$ associated costs are unit production cost $p_{t}$, setup cost $f_{t}$, and unit inventory holding cost $h_{t}$. Note that setup cost $f_{t}$ is not per unit based. For each period $t \in \left\{1,2,\ldots,T \right\}$ production capacity is denoted by $c_{t}$. Without loss of generality, all parameters can be assumed to be non-negative. The number of units produced and ending inventory in period $t$ is represented by non-negative variables $x_{t}$ and $s_{t}$, respectively. Binary variable $y_{t}$ takes value 1 if there is production in period $t$, and takes value 0 otherwise. The CLSP can be formulated as:
\begin{subequations}
\label{clsp:1}
\begin{eqnarray}
\min &&     \sum_{t=1}^{T} (p_{t}x_{t}  + f_{t}y_{t} + h_{t}s_{t})\label{objective1-ex:1}\\
\textrm{s.t.} &&
      s_{t-1} + x_{t} - d_{t} = s_{t} \quad \quad \ \forall t=1,2,\ldots,T \label{cons1-ex:1}\\
      && x_{t} \leq y_{t}c_{t} \quad \quad \ \forall t=1,2,\ldots,T \label{cons2-ex:1}\\
      && x_{t},s_{t} \geq 0 \quad \quad \ \forall t=1,2,\ldots,T \label{nonneg1-ex:1}\\
     && y_{t} \in \{0,1\} \quad \quad \quad  \ \forall t=1,2,\ldots,T. \label{integrality1-e:1}
\end{eqnarray}
\end{subequations}
The objective function \eqref{objective1-ex:1} minimizes the sum of production costs, setup costs, and inventory holding costs over all periods $t \in \left\{1,2,\ldots,T \right\}$. Constraints \eqref{cons1-ex:1} ensure the inventory flow over multiple periods. Specifically, the demand in period $t$ must be satisfied by inventory at the end of period $t-1$ and units produced in period $t$. The remaining amount is the inventory at the end of period $t$. Constraints \eqref{cons2-ex:1} limit the production by capacity and ensure that a fixed cost of production is incurred in the objective function if there is production in period $t$. Constraints \eqref{nonneg1-ex:1} enforce that the amounts of units produced and kept in inventory are non-negative. Finally, constraints \eqref{integrality1-e:1} ensure that $y_t$ are binary variables. The parameter $s_{0}$ represents the initial inventory and is assumed to be zero.

\section{LSTM-Optimization Framework}	
\label{LSTM Framework}

In this section, we present the LSTM-Opt framework that we develop to predict the optimal solution of the CLSP. Using the LSTM-Opt framework, we only predict the binary decision variables $y_{t}$ that correspond to a production decision instead of predicting all decision variables. As depicted in Figure \ref{LSTM-Opt}, the LSTM-based framework starts with data generation. The datasets with different characteristics are constructed according to the data-generation scheme described in Section \ref{CLSP Instance Generation}. The resulting datasets are divided into three categories involving the training, validation, and test sets, which consist of 64\%, 16\%, and 20\% of the data, respectively. The LSTM network parameters are optimized using a training set. This is done by minimizing a loss function that measures the performance of the model's predictions compared to actual values. The binary cross-entropy is a common choice as a smooth loss function for binary classification because it leads to faster training with a better generalization performance than the sum of squares error \citep{bishop1995neural}. The objective function of the CLSP given by equation \eqref{objective1-ex:1} is not minimized by the LSTM network. In the training step, we minimize the binary cross-entropy loss function given in the following equation
\eqref{lossfct}:

\begin{figure}[!tb]

   \centering
   \caption{LSTM-Opt framework.}\label{LSTM-Opt}
   \scalebox{0.9}{\includegraphics[width=\linewidth]{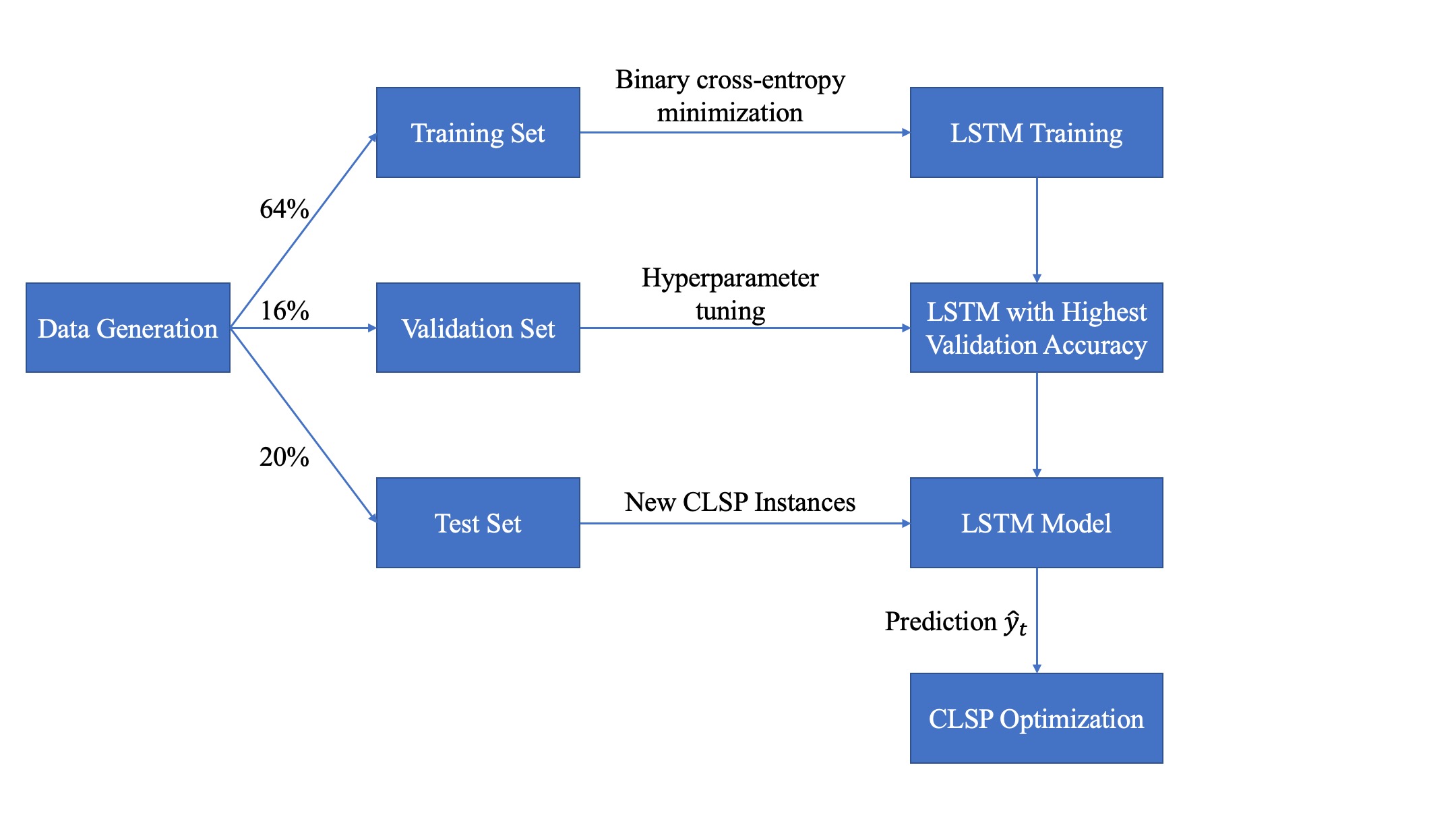}}
   
\end{figure}

 \begin{equation}
\label{lossfct}
\L(y^*,\hat{y}) = -\frac{1}{T} \times \sum_{t=1}^{T} (y^*_{t} \times log(\hat{y}_{t}) + (1-y^*_{t}) \times log(1-\hat{y}_{t})) 
\end{equation}
where $y^*$ represents the optimal values of the binary decision variables, and $\hat{y}$ represents predicted values of the binary decision variables of a CLSP instance. The binary cross-entropy loss function in equation \eqref{lossfct} measures the discrepancy between $y^*$ and $\hat{y}$. 

\begin{figure}[!htb]

  \centering
     \caption{Bidirectional LSTM \citep{schuster1997bidirectional} adapted to represent the CLSP multi-period structure.}\label{LSTMstructure}
   \scalebox{0.90}{\includegraphics[width=\linewidth]{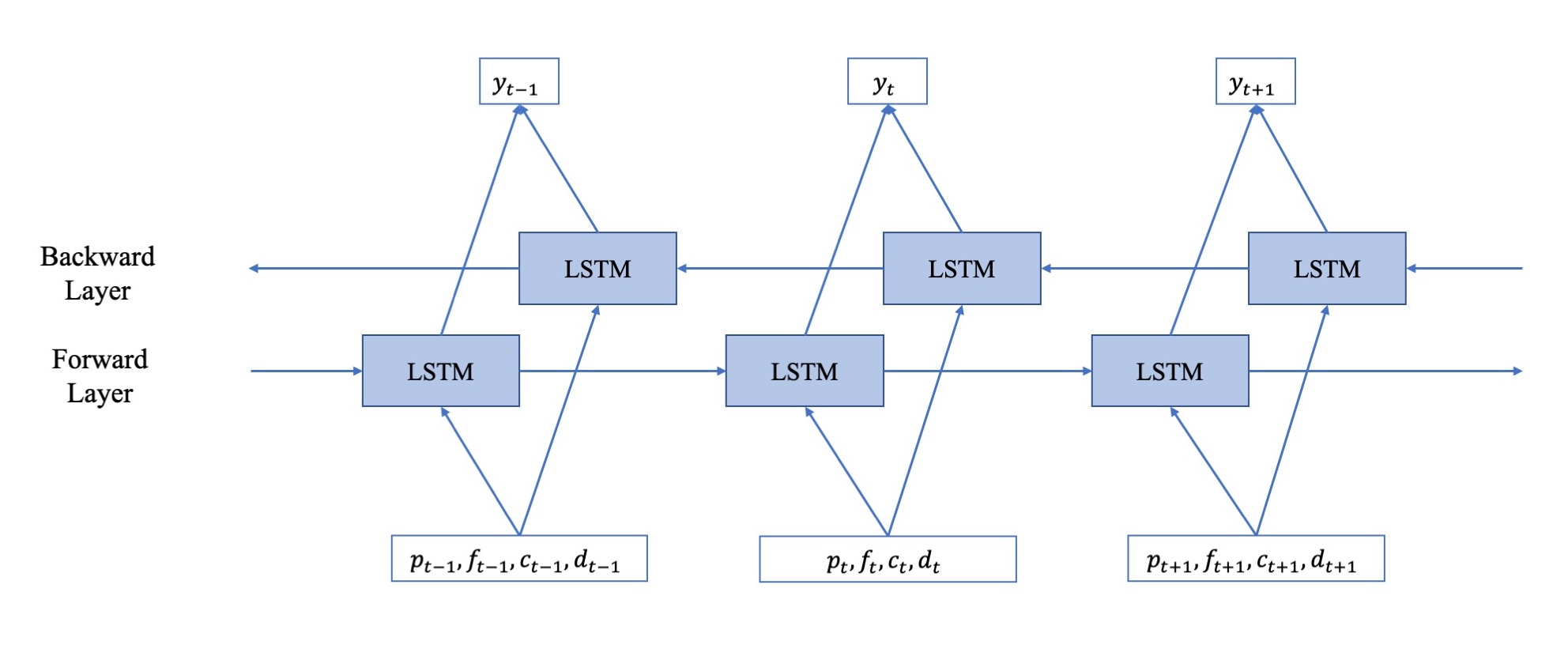}}
\end{figure}

The LSTM model consists of several bidirectional LSTM layers that can process the information in both time directions and an output layer with a sigmoid activation function. Figure \ref{LSTMstructure} shows the flow of information in the forward and backward layers in bidirectional LSTM. The input layer for LSTM consists of available features for that period: unit production cost $p_{t}$, setup cost $f_{t}$, production capacity $c_{t}$, and demand $d_{t}$ for $t \in \left\{1,2,\ldots,T \right\}$. Note that we omitted the holding cost $h_{t}$ because it is taken as constant. For period $t$, information is carried from period $t-1$ in the forward layer and used to generate output in period $t$. In the backward layer, information is carried from period $t+1$ to period $t$, and it is used to generate output together with inputs in period $t$. The outputs of forward and backward layers are combined to generate prediction $\hat{y}_{t}$. After each hidden layer, a dropout layer is added for regularization.

We compare the models with different parameters, using the instances in the validation set, in a method known as hyperparameter tuning. We then choose the model with the highest validation accuracy, which is the proportion of the correctly predicted variables. Note that the validation set is not used to minimize the binary cross-entropy in equation \eqref{lossfct}; it is only used to compare LSTM networks with different hyperparameters, such as learning rate, number of layers, hidden nodes, and dropout rate. Then for each instance in the test set, a prediction is generated using the picked model. The framework described does not provide results on the feasibility of the resulting prediction and how good it is compared to the objective function value. The resulting predictions are added to problem \eqref{clsp:1} as constraints, and then CLSP is re-solved using CPLEX. The described approach can deliver optimal solutions fast and accurately without much loss in feasibility and optimality, as demonstrated in the computational results under Section \ref{Results}.

CLSP is an MIP because of the binary decision variables. Predicting all binary decision variables and then fixing the predicted values in the MIP formulation \eqref{clsp:1} makes the problem a linear program, which yields a significant reduction in the solution time. This approach often leads to infeasibility due to its strict nature. Instead, predicting some of the binary decision variables results in more flexibility when resolving the problem instance and reduces the number of infeasible problems while still improving the solution time.

Additionally, the integral nature of the other two decision variables is preserved by solely predicting the binary variable because once the binary variables are fixed in the CLSP, it reduces to a linear program \citep{pochet2006production}. Also, predicting the binary variable carries an interpretable meaning of the production decision and its timing. Once the decision of whether to produce or not is determined and fixed in a period, the MIP solver determines the amount of production and the inventory levels.

Our integrated ML+OR tool can be beneficial for real-time applications where problems with different parameters are solved repeatedly. Lot-sizing and its variants commonly arise in the energy, pharmaceutical, electronics, food, processing, and consumer goods industries \citep{copil2017simultaneous}. After an ML model is trained, it is not necessary to update the trained model after each prediction. Therefore, once an ML model is trained, predictions can be achieved in milliseconds by an LSTM forward pass to solve many CLSP instances in a quite fast manner.

\section{Implementation and Experimentation}	
\label{Implementation and Experimentation}	
This section presents the CLSP instance generation scheme and the implementation details of our LSTM-Opt framework. All the codes are written in C++ and Python to generate CLSP instances and run the LSTM-Opt framework. 

\subsection{CLSP Instance Generation}
\label{CLSP Instance Generation}

The training, validation, and testing data were generated by the scheme presented in \citet{atamturk2004a}. The difficulty of problems was determined by two main factors: tightness of the capacities with respect to demand and the ratio between setup and holding cost. Following the parameters used in \citet{buyuktahtakin2016dynamic}, instances are generated from capacity-to-demand ratios $c \in \left\{3,5,8\right\}$, setup-to-holding cost ratios $f \in \left\{1,000,10,000\right\}$ and the number of periods $T \in \left\{90,120\right\}$. The parameters regarding demand $d_{t}$, unit production cost $p_{t}$, production capacity $c_{t}$, and setup cost $f_{t}$ are generated from integer uniform distribution with the ranges $d_{t} \in \left[1,600\right]$, $p_{t} \in \left[1,5\right]$, $c_{t} \in \left[0.7c\bar{d},1.1c\bar{d}\right]$, $f_{t} \in \left[0.9f\bar{h},1.1f\bar{h}\right]$, where $\bar{d}= \frac{1}{T} \left(\sum_{t=1}^{T} d_{t}\right)$ and $\bar{h}= \frac{1}{T} \left(\sum_{t=1}^{T} h_{t}\right)$, respectively. Unit inventory holding cost $h_{t}$ is set at one at each period.

For each of the 12 combinations of parameters $c, f$, and $T$ as described above, 100,000 instances (problems) are generated, resulting in a total of 1,200,000 instances. All instances are solved using CPLEX. Infeasible problems are eliminated and replaced by feasible instances by regenerating new instances. The training, validation, and test set consists of 64,000, 16,000, and 20,000 CLSP instances, respectively, for each combination of parameters. 
 
\subsection{LSTM-Opt Implementation Details }
\label{trainsection}

Before the training, the data is standardized by subtracting the feature mean and dividing by feature standard deviation as a preprocessing step, which is often practically useful for faster convergence if different inputs have typical values that differ significantly \citep{lecun2012efficient}. In the hyperparameter tuning step, we compared LSTM models with different parameters, such as learning rate, number of layers, hidden nodes, and dropout rate using the validation set. The values considered are $ \lbrack 2, 6\rbrack$ for the number of hidden layers, $ \lbrack 10, 150\rbrack$ for the number of units in hidden layers, $ \lbrack 0.1, 0.5\rbrack$ for the dropout rate, and $ \lbrack 0.1, 0.001\rbrack$ for the learning rate. The selected LSTM model contains three hidden LSTM layers, each with 40 hidden units in each time direction. Therefore, bidirectional LSTM for each layer has 80 hidden units. After each LSTM layer, a dropout layer with a drop rate of 0.3 is added to regularize the network. We used Adam optimizer with an initial learning rate of 0.01, which is an adaptive learning rate optimization algorithm that has been shown to work well in practice \citep{kingma2014adam}.

For each instance in the test set, the values of binary variables are predicted. For each period $t \in \left\{1,2,\ldots,T \right\}$, the LSTM network generates a prediction in the range of $ \lbrack0, 1\rbrack$ for each $y_{t}$. The value of $\max(\hat{y_{t}},1-\hat{y_{t}})$ for $t \in \left\{1,2,\ldots,T \right\}$, where $\hat{y_{t}}$ represents the predicted value of the binary variable $y_{t}$, is calculated and ordered in decreasing order. Predicted variables are selected up to the desired level, and $\hat{y_{t}}$ is labeled as 0 or 1 using a cut-off value of 0.5. Those variables can be interpreted as the ones closest to either zero or one, and thus we are more confident in the LSTM model’s prediction. Let $D \subseteq T$ be the set of indices of those binary decision variables predicted and selected using the $\max(\hat{y_{t}},1-\hat{y_{t}})$ function and a pre-set prediction percentage. Finally, those values are added as a constraint to the original model \eqref{clsp:1}, as shown in the following modified problem \eqref{clsp:2}:
\begin{subequations}
\label{clsp:2}
\begin{eqnarray}
\min &&     \sum_{t=1}^{T} (p_{t}x_{t}  + f_{t}y_{t} + h_{t}s_{t})\label{objective1-ex:2}\\
\textrm{s.t.} && Constraints \quad \eqref{cons1-ex:1}-\eqref{integrality1-e:1}\label{prevcons}\\
 && y_{t} = \hat{y_{t}} \quad \quad \quad  \ \forall t\in D . \label{pred-e:2}     
\end{eqnarray}
\end{subequations}
Problem \eqref{clsp:2} is solved again to assess the quality of the LSTM predictions.

\section{Computational Results}
\label{Results}
This section presents results from computational experiments performed using the LSTM-Opt framework described in Section \ref{LSTM Framework} on randomly generated CLSP instances with various characteristics as defined in Section \ref{Implementation and Experimentation}. All experiments are performed on a computer running Windows 10 Intel i7 with 3.6 GHz GPU and 64 GB of memory. The CLSP instances are solved with IBM ILOG CPLEX 12.7.1. All of the results regarding the test-set solution times are presented in CPU seconds. The detailed results for training LSTM models are presented in Appendix \ref{Results for Training LSTM Models}.

We solve problem \eqref{clsp:1} instances using the default CPLEX as a benchmark to compare the performance of our LSTM-Opt framework for solving similar and different CLSP instances. The test dataset consists of 20,000 CLSP instances for each combination of the parameters $f$, $T$, and $c$. 

As an alternative to solving the problem \eqref{clsp:2}, where we fix the predicted values in the CLSP formulation \eqref{clsp:1} as constraints, we utilize two CPLEX solver methods--AddUserCuts and AddMIPStart methods of CPLEX to eliminate infeasibility as described below:

\begin{itemize}
\item \textbf{100(UC)}: AddUserCuts method of CPLEX, which enable CPLEX to add cuts into the user cut pool and use the cuts as needed.
\item \textbf{100(MS)}: AddMIPStart method of CPLEX, which enable CPLEX to provide a starting solution to the model with a user cut pool.
\end{itemize}


\subsection{Quality of Predictions}
\label{Quality of Predictions}

Here, we present a number of metrics with their formal definitions that are used to evaluate the effectiveness of our LSTM-Opt framework. Specifically, those metrics assess the proposed method to solve optimization instances with respect to the improvement in the solution time as well as the feasibility and optimality of the resulting solutions. The following metrics are used in Tables \ref{t:f10,000t120}-\ref{gen4extra} and \ref{t:f1,000t90}-\ref{gen1}:

\begin{itemize}
\item \textbf{timeCPX}: Mean solution time of a CLSP instance of the problem \eqref{clsp:1} in CPU seconds without any predictions using default CPLEX.
\item \textbf{timeML}: Mean solution time of the LSTM-Opt framework, including the prediction generation time by the LSTM model.
\item \textbf{pred(\%)}: Percent of binary variables predicted by the LSTM-Opt framework.
\end{itemize}

Additionally, we provide the following metrics and their formal definitions and use a combination of them to present the results:

\begin{definition}
\label{improvement}
The solution time factor improvement factor obtained by fixing the predicted variables as a constraint (or using AddUserCuts and addMIPStart) is given by:
\begin{equation}
\label{equationimprovement}
\mathbf{timeimp} = \frac{timeCPX}{timeML}.
\end{equation}
\end{definition}

\begin{definition}
\label{timegain}
The percent solution time gain obtained by fixing the predicted variables as a constraint (or using AddUserCuts and addMIPStart) is given by:
\begin{equation}
\label{equationtimegain}
\mathbf{timegain (\%)} = \frac{(timeCPX-timeML)}{timeCPX} \times100.
\end{equation}
\end{definition}

\begin{definition}
\label{infeasibility}
The percent infeasibility of a test set resulted from using predicted binary variables is given by:
\begin{equation}
\label{equationinfeasibility}
\mathbf{inf (\%)} = \frac{\hat{m}}{m} \times100,
\end{equation}
where $\hat{m}$ represents the number of CLSP instances that become infeasible by adding predictions as a constraint and $m$ represents the total number of CLSP instances in the test set.
\end{definition}

\begin{definition}
\label{optgapdef}
Let $\left(x^*,y^*,s^*\right)$ be the optimal solution for the original MIP problem \eqref{clsp:1} that is obtained by the CPLEX solver and ${Z}(x^*,y^*,s^*)$ be the corresponding optimal objective value. Let $\hat{y}$ be the partial or full prediction of binary variables, $\left(\tilde{x},\tilde{y},\tilde{s}\right)$ be the optimal solution obtained by CPLEX using predictions $\hat{y}$ in problem \eqref{clsp:2}, and ${Z}(\tilde{x},\tilde{y},\tilde{s})$ be the resulting objective function value. Note $\tilde{y}$ is equivalent to $\hat{y}$ when a full prediction is made. The optimality gap due to using solutions in our LSTM-Opt prediction framework is defined over feasibly-solved instances as follows:

\begin{equation} 
\label{equationoptgap}
\mathbf{optgap (\%)}  =   \frac{\left({Z}(\tilde{x},\tilde{y},\tilde{s})-{Z}(x^*,y^*,s^*)\right)}{{Z}(x^*,y^*,s^*) } \times 100.
\end{equation} 
\end{definition}

In the next section, we present computational results to demonstrate the effectiveness of our prediction-optimization method, using the training and test instances with the same distribution. The predictions fed into the CPLEX solver may not be feasible for the CLSP instance. The test set instances for which the LSTM prediction leads to an infeasible solution are not included in the calculations of timeML, timeimp, timegain(\%), and optgap(\%).

\subsection{Predicting Instances with Same Distribution}
Each row of Tables \ref{t:f10,000t120}, \ref{gen2}, \ref{t:f1,000t90}, \ref{t:f10,000t90}, \ref{t:f1,000t120}, and \ref{gen1} presents the averages of 20,000 instances, whereas Tables \ref{gen3}, \ref{gen4}, and \ref{gen4extra} present the mean result for 10 instances due to long solution times for each $c$ value and each $f-T$ pair. Table \ref{t:f10,000t120} presents results for instances with $T = 120$ with $f = 10,000$. The dataset of $c = 3$ is harder than the dataset with $c=5$ and $c=8$. Both 25\% and 50\% prediction levels achieve all-feasible predictions that do not increase the objective function value. With the 50\% prediction level, the mean solution time decreases by more than 10-fold. As the prediction level increases, the time factor improvement increases as well. Predictions at the 75\% level reduce the solution time by a factor of 50 with an infeasibility of the test set below 0.3\% and without any optimality gap. However, as the prediction levels increases, predictions lead to more infeasible instances in the test set. At the full prediction level (pred(\%)=100), more than half of the predictions results in infeasible solutions to problem \eqref{clsp:2}; however, the issue of infeasibility is remedied by the CPLEX’s user cuts approach (AddUserCuts), which eliminates the infeasible solutions in the cut pool. The user cuts (UC) approach provides a significant solution time factor improvement of 68 with an optimality gap of over 1\% without any infeasibility in the test set. The detailed results of experiments with $f = 10,000$, $T = 90$ and $f = 1,000$, $T = 90,120$ are presented in Tables \ref{t:f1,000t90}-\ref{t:f1,000t120} in Appendix \ref{More Results on the Experiments}.

\renewcommand{\baselinestretch}{1}
\begin{table}[!htb]
\caption{Summary of experiments for $f = 10,000$ and $T = 120$}
\label{t:f10,000t120}
\centering
\scalebox{0.95}{
\begin{tabular}{cccccccc}
$c$     	&	       pred(\%)        	&	timeCPX	&	timeML	&	timeimp	&	timegain(\%)	&	inf(\%)	&	optgap(\%)	\\	﻿\hline
3	&	25	&	22.6	&	6.9	&	3	&	69.3	&	0.0	&	0.0	\\	
        	&	50	&		&	1.7	&	13	&	92.5	&	0.0	&	0.0	\\	
        	&	75	&		&	0.4	&	50	&	98.0	&	0.3	&	0.0	\\	
        	&	85	&		&	0.3	&	84	&	98.8	&	1.7	&	0.1	\\	
        	&	90	&		&	0.2	&	94	&	98.9	&	4.3	&	0.3	\\	
        	&	95	&		&	0.2	&	104	&	99.0	&	13.5	&	0.8	\\	
        	&	100	&		&	0.1	&	208	&	99.5	&	57.5	&	2.1	\\	
        	&	       100(MS) 	&		&	0.3	&	68	&	98.5	&	0.0	&	1.3	\\	
        	&	       100(UC) 	&		&	0.3	&	68	&	98.5	&	0.0	&	1.3	\\	\hline
5	&	25	&	3.0	&	2.1	&	1	&	28.7	&	0.0	&	0.0	\\	
        	&	50	&		&	1.3	&	2	&	56.0	&	0.0	&	0.0	\\	
        	&	75	&		&	0.5	&	6	&	83.5	&	0.0	&	0.0	\\	
        	&	85	&		&	0.3	&	9	&	88.6	&	0.3	&	0.0	\\	
        	&	90	&		&	0.3	&	11	&	90.7	&	0.8	&	0.1	\\	
        	&	95	&		&	0.2	&	12	&	91.6	&	3.0	&	0.2	\\	
        	&	100	&		&	0.1	&	33	&	97.0	&	24.0	&	1.7	\\	
        	&	       100(MS) 	&		&	2.8	&	1	&	8.4	&	0.0	&	0.0	\\	
        	&	       100(UC) 	&		&	0.3	&	9	&	89.2	&	0.0	&	0.7	\\	\hline
8	&	25	&	1.4	&	1.0	&	1	&	29.4	&	0.0	&	0.0	\\	
        	&	50	&		&	0.6	&	2	&	56.7	&	0.0	&	0.0	\\	
        	&	75	&		&	0.4	&	3	&	69.9	&	0.0	&	0.0	\\	
        	&	85	&		&	0.4	&	4	&	72.7	&	0.1	&	0.0	\\	
        	&	90	&		&	0.3	&	4	&	77.5	&	0.1	&	0.0	\\	
        	&	95	&		&	0.3	&	5	&	81.3	&	0.4	&	0.0	\\	
        	&	100	&		&	0.1	&	17	&	94.1	&	5.4	&	1.1	\\	
        	&	       100(MS) 	&		&	1.3	&	1	&	10.1	&	0.0	&	0.0	\\	
        	&	       100(UC) 	&		&	0.3	&	4	&	77.6	&	0.0	&	0.4	\\	\hline

\end{tabular}}
\end{table}
\renewcommand{\baselinestretch}{}

Figures \ref{figT90f1000}-\ref{figT120f10000} summarize the results with the optgap(\%), inf(\%), and timeimp for changing c for instances with $T = 90,120$ and $f = 1,000,10,000$. Figures \ref{figT90f1000}-\ref{figT120f10000} show that as the level of predicted variables increases, the time improvement also increases at the price of an increased optimality gap and infeasibility in the test set. The problems are harder when $c = 3$ ($f = 10,000$) compared to $c = 8$ ($f = 1,000$) for the same $T$. Predicting at lower levels provides good results for harder problems with significant time improvement without causing much optimality gap and infeasibility, e.g., a time improvement factor of 13 is achieved with the 50\% prediction level without any infeasibility or optimality gap for instances with $c = 3$, $T = 120$, and $f = 10,000$ (Figure \ref{figT120f10000}). The time improvement factor is the highest when using the 100\% prediction. For instances with $f=1,000$, the full (100\%) prediction results in less than a 1.5\% infeasibility. However, it could provide high levels of infeasibility in the test set for harder problems with $f=10,000$. On the other hand, the 50\% prediction level provides over a time factor improvement of 3 and reduces the infeasibility to 0.01\% and the optimality gap to zero (Figure \ref{figT90f10000}). When $f = 1,000$, time improvement increases significantly with the level of prediction, but the increase in the optimality gap and infeasibility is much less than the counterpart instances with $f = 10,000$, e.g., an infeasibility of 0.4\% and optimality gap of zero is obtained for instances with $c = 5$, $T = 90$, and $f = 1,000$ compared to an infeasibility of 23.3\% and optimality gap of 1.2\% is observed for instances with $c = 5$, $T = 90$, and $f = 10,000$, using full predictions.

\begin{figure}[!htb]
\centering
\caption{Summary of Results with optgap(\%), inf(\%), and timeimp}\label{figresults}
\addtocounter{figure}{-1}
\subfloat[$T = 90$, $f = 1,000$\label{figT90f1000}]{%
  \includegraphics[width=0.44\textwidth]{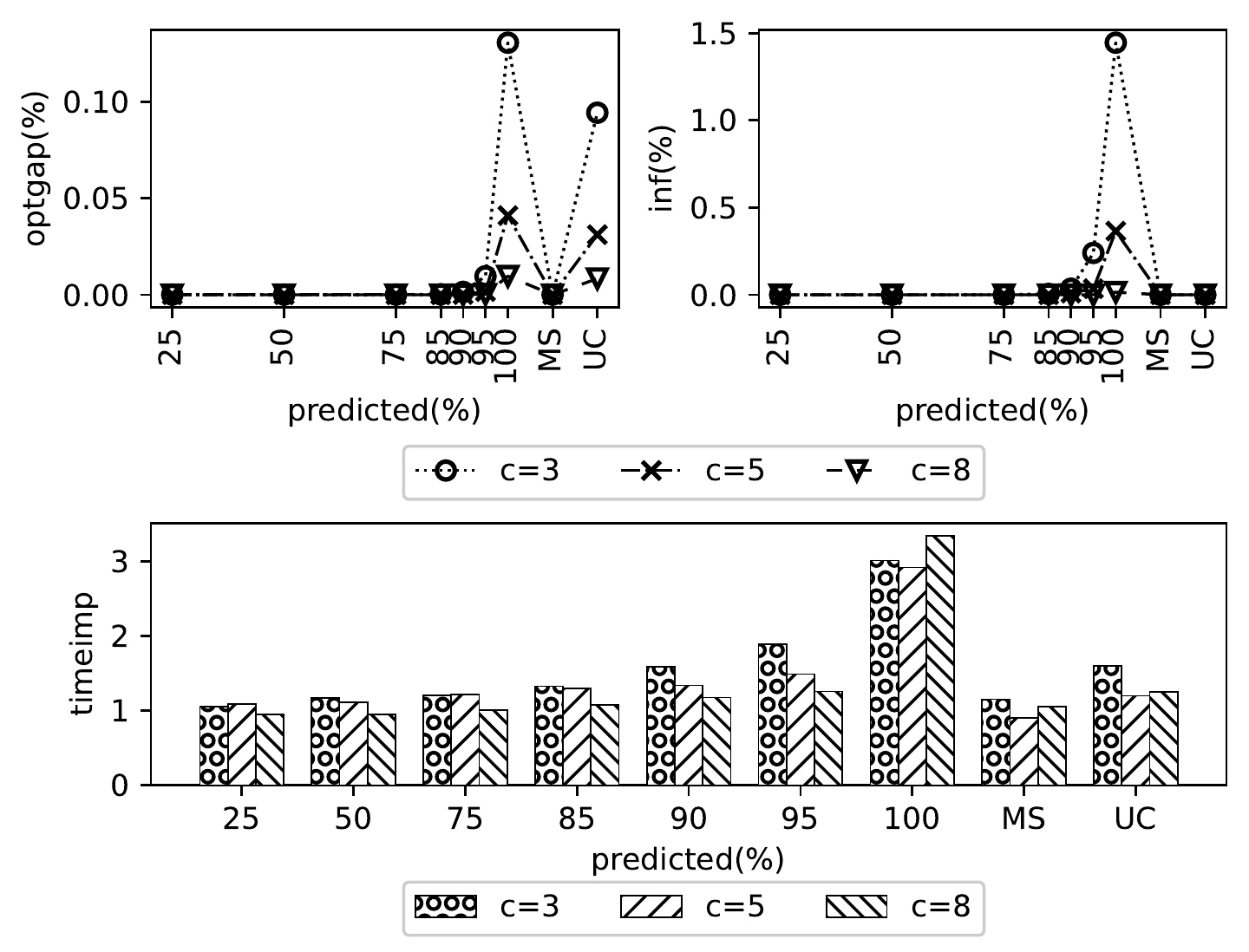}
}%
\subfloat[$T = 90$, $f = 10,000$\label{figT90f10000}]{%
  \includegraphics[width=0.44\textwidth]{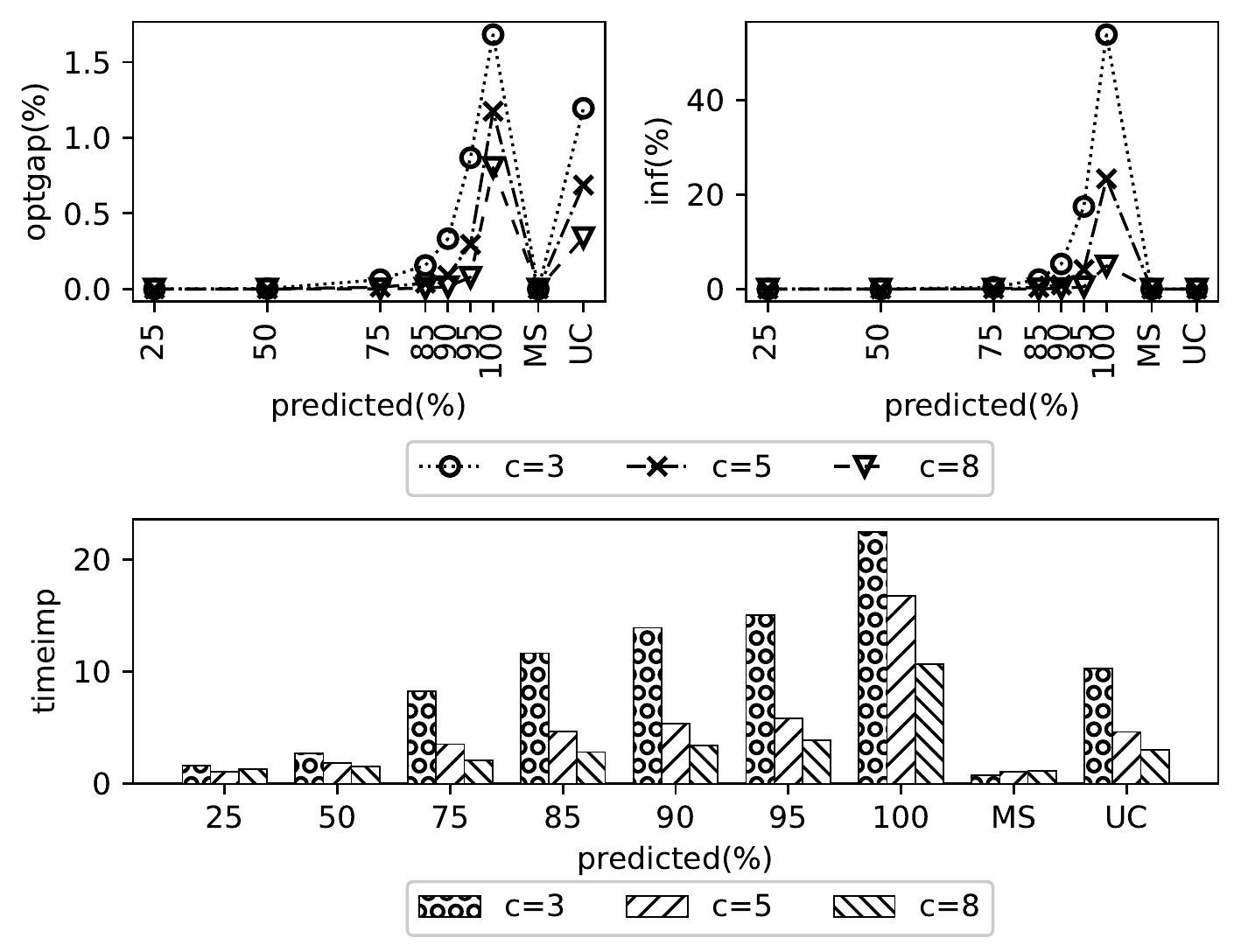} 
}

\subfloat[$T = 120$, $f = 1,000$\label{figT120f1000}]{%
  \includegraphics[width=0.44\textwidth]{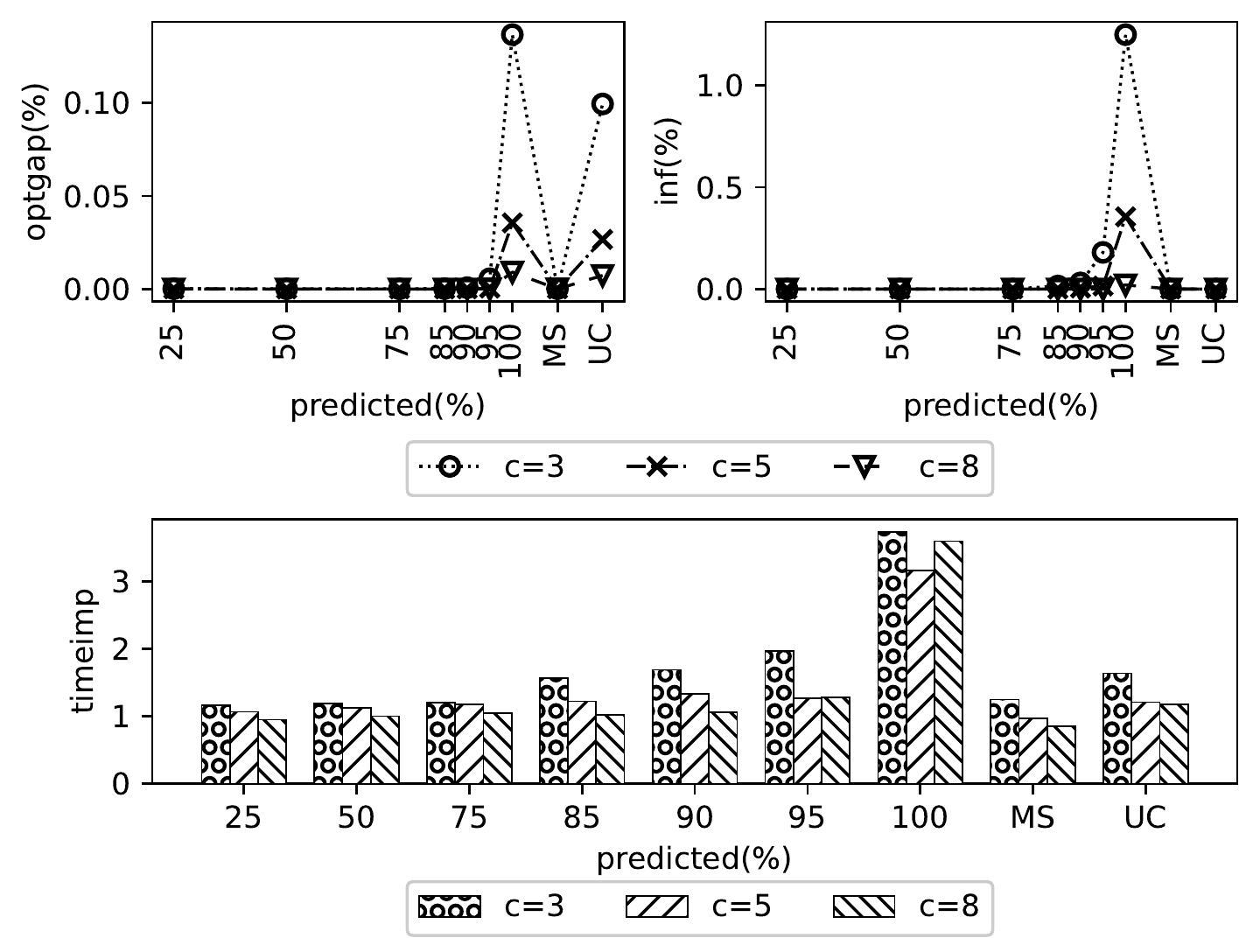}%

}%
\subfloat[$T = 120$, $f = 10,000$\label{figT120f10000}]{%
  \includegraphics[width=0.44\textwidth]{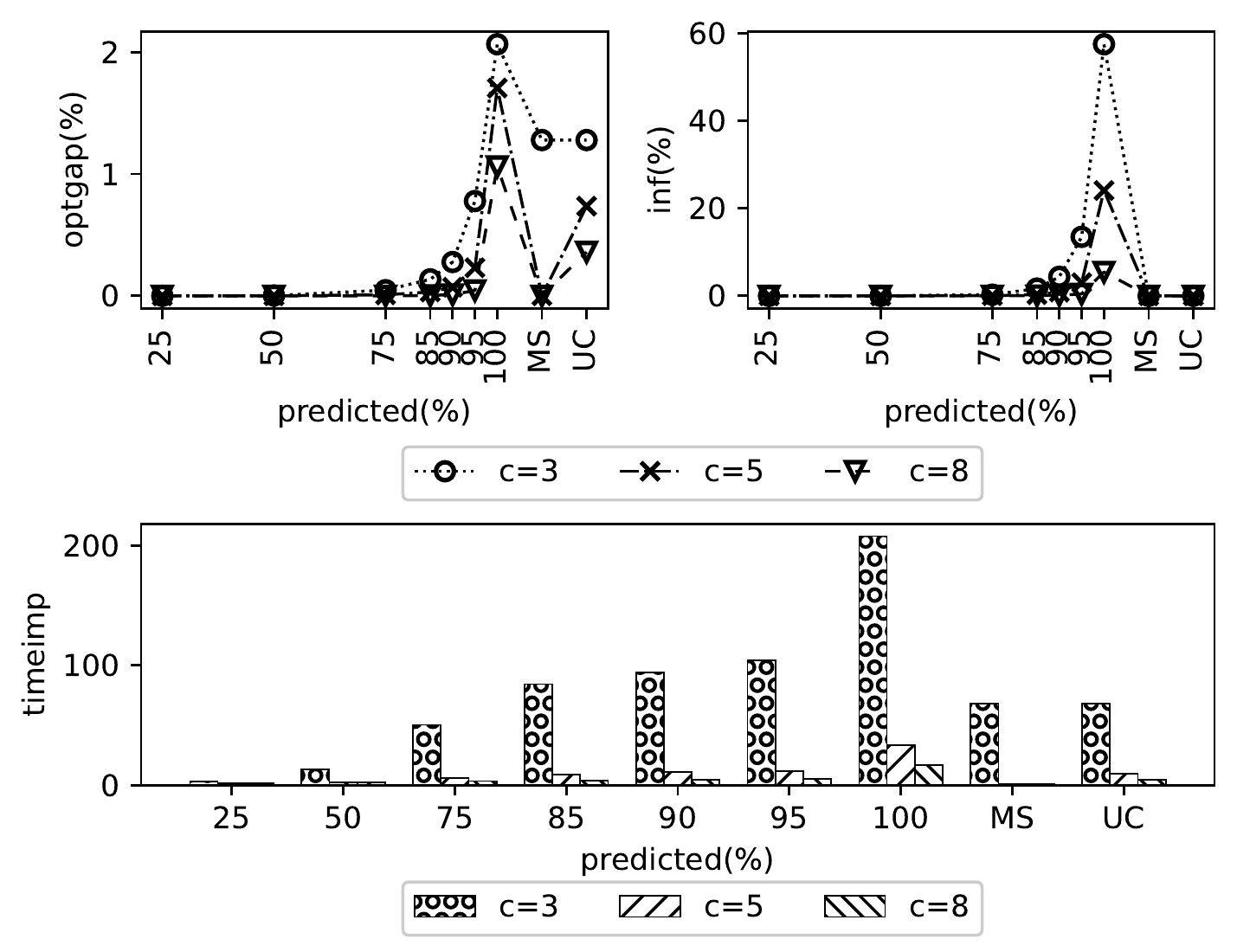}%

}
\stepcounter{figure}

\end{figure}

Figures \ref{figcavg}-\ref{figallavg} show averages for changing $c$, $f$, and $T$, and the overall average. As the value of $c$ increases, the optimality gap, infeasibility, and time improvement generally decrease (Figure \ref{figcavg}). Figure \ref{figfavg} shows that the value of $f$ has a significant impact on results. The optimality gap and infeasibility are significantly lower when $f = 10,000$, with lower-level predictions. Also, the time factor improvement is significantly greater at all prediction levels when $f = 10,000$. Both the optimality gap and time improvement are slightly higher when $T = 120$ compared to the instances with $T = 90$. Figure \ref{figallavg} shows that using a prediction level of around 85\% can balance all evaluation metrics by providing a solution time factor improvement of 9.

\begin{figure}[!htb]
\centering
\caption{Summary of Results with Different Data Generation Parameters}\label{figresults2}
\addtocounter{figure}{-1}
\subfloat[Averages for $c$\label{figcavg}]{%
  \includegraphics[width=0.44\textwidth]{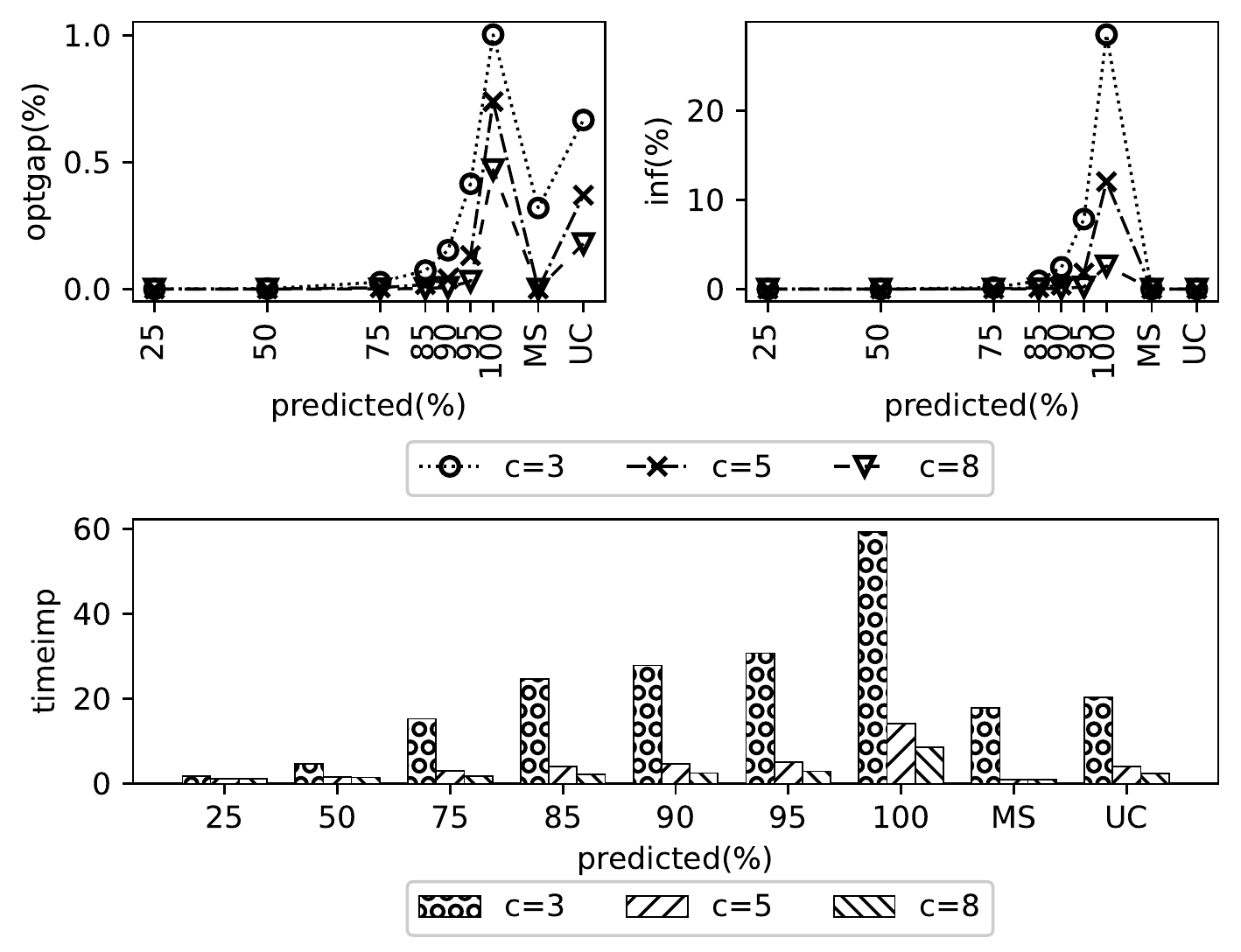}%
}%
\subfloat[Averages for $f$\label{figfavg}]{%
  \includegraphics[width=0.44\textwidth]{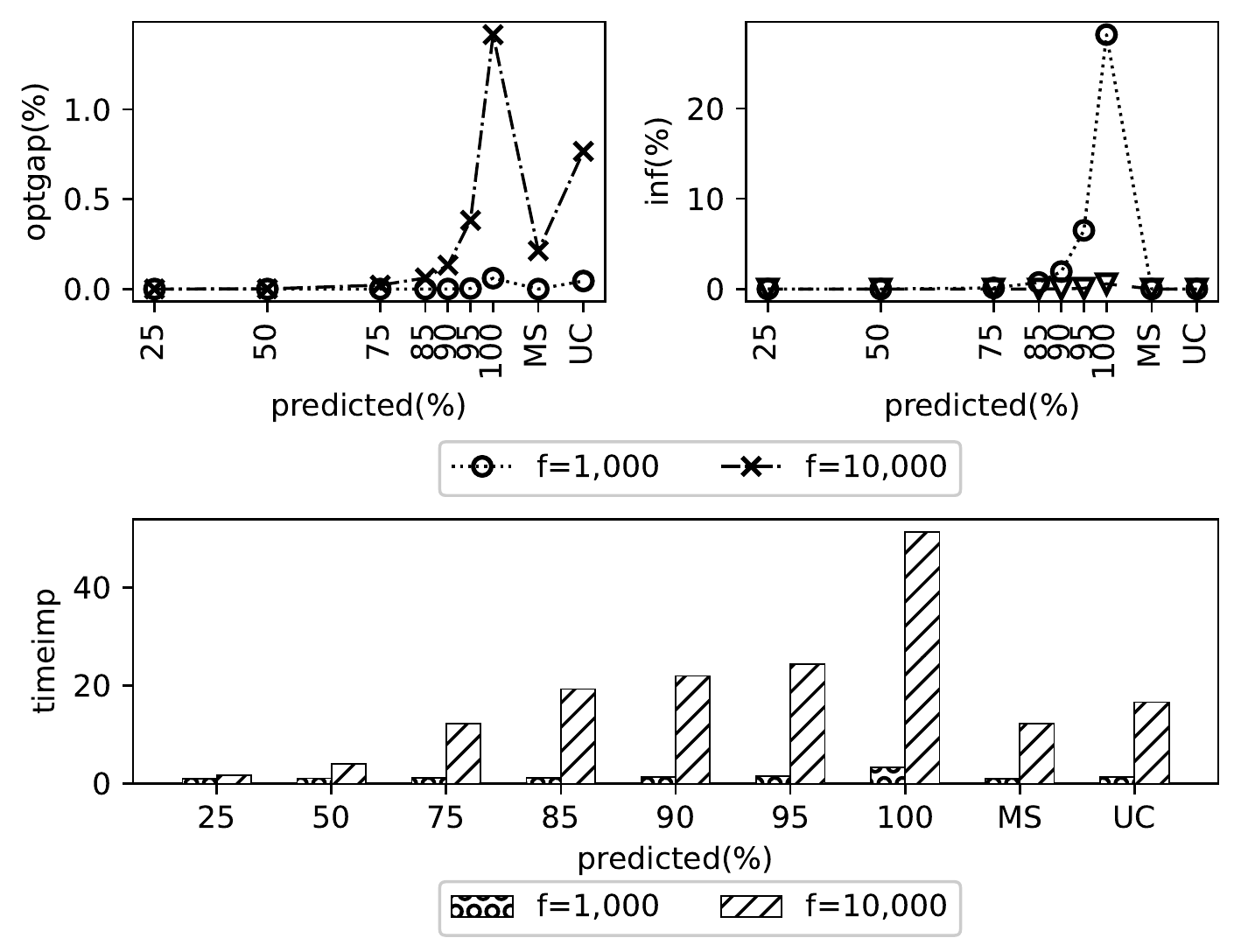}%
}

\subfloat[Averages for $T$\label{figTavg}]{%
  \includegraphics[width=0.44\textwidth]{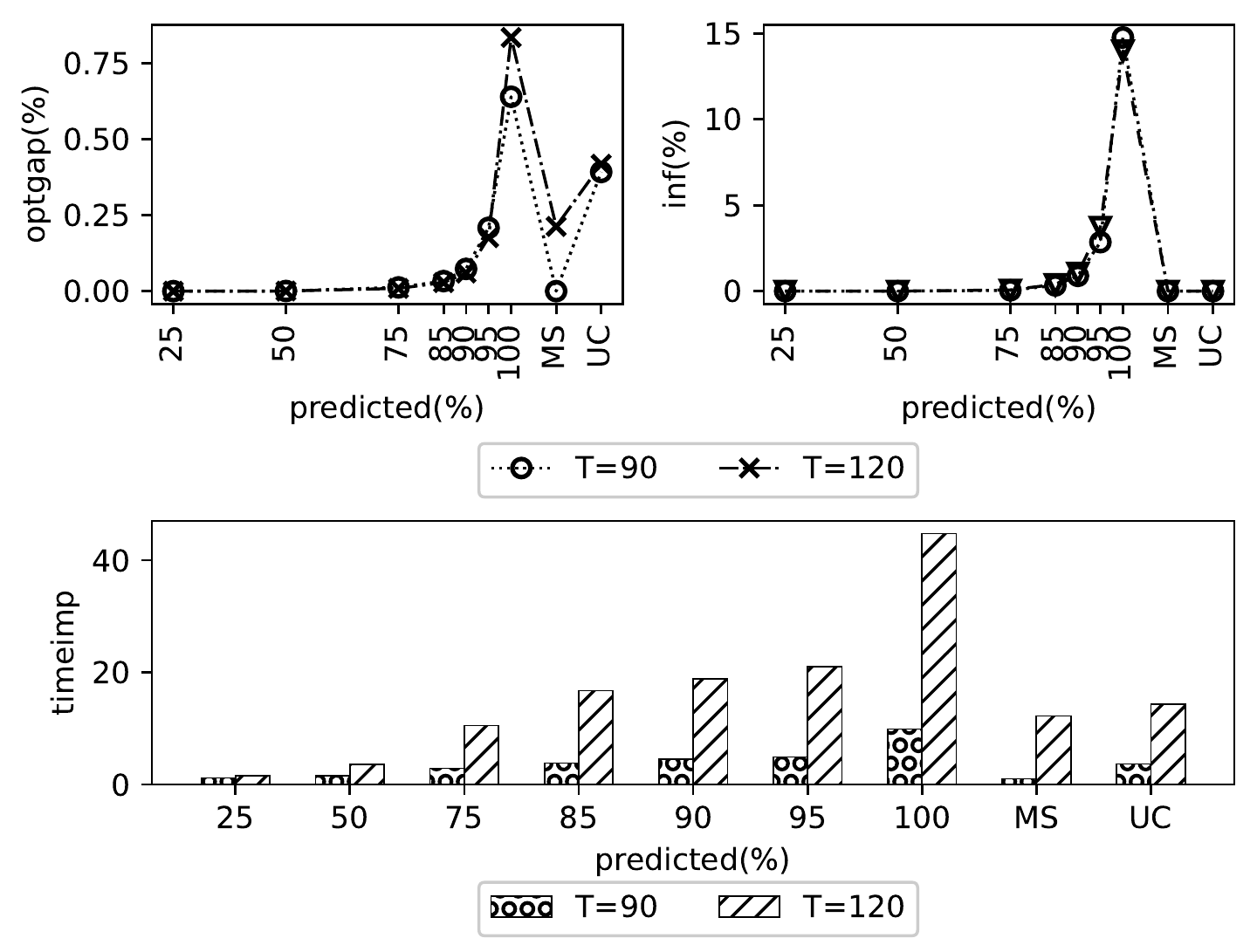}%
}%
\subfloat[Overall averages\label{figallavg}]{%
  \includegraphics[width=0.44\textwidth]{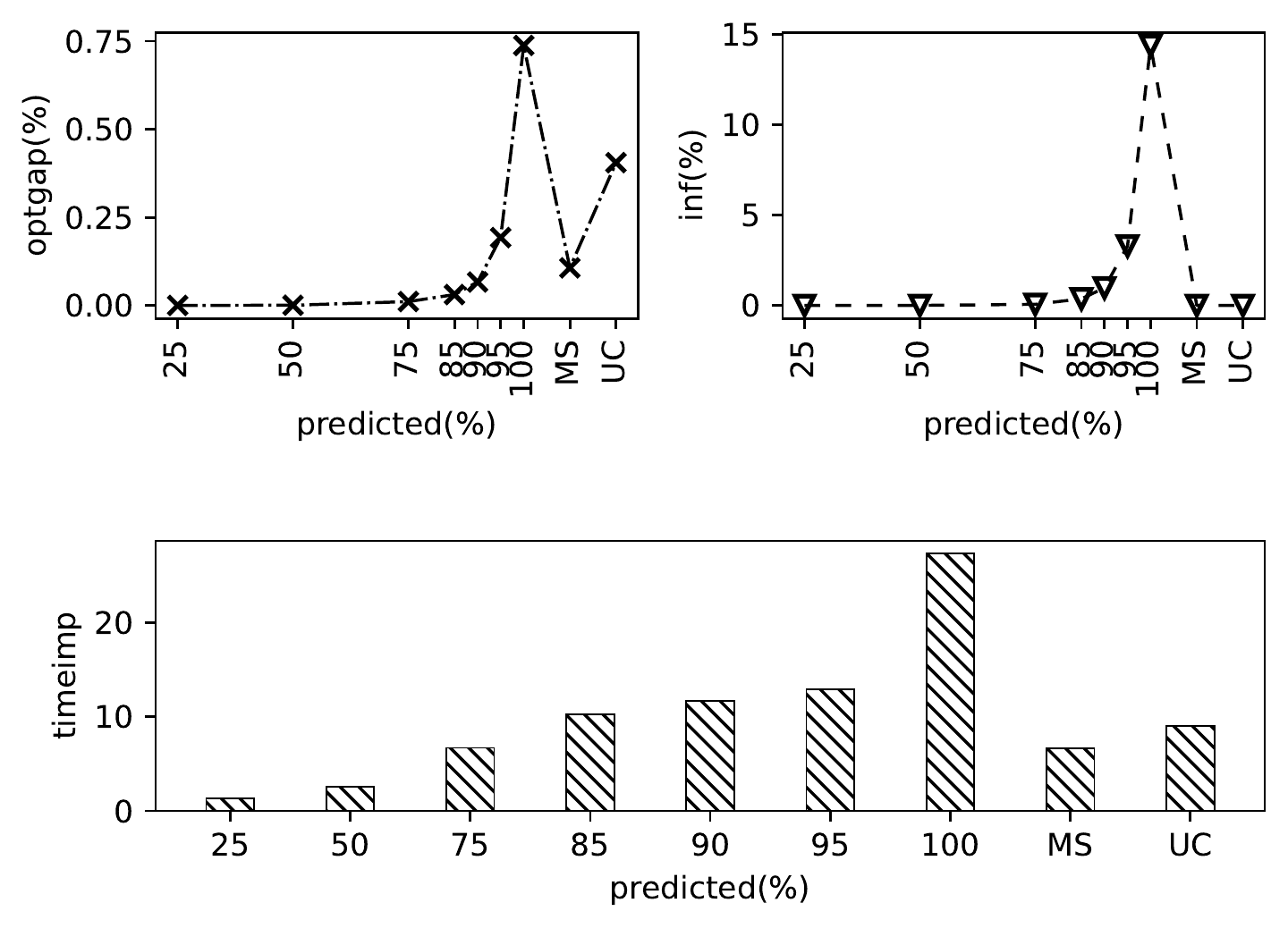}%
}

\stepcounter{figure}

\end{figure}

Table \ref{t:average} shows the averages presented in Tables \ref{t:f10,000t120}, \ref{t:f1,000t90}, \ref{t:f10,000t90}, and \ref{t:f1,000t120} for the LSTM-Opt 85\% prediction level, the 100(MS), and the 100(UC). When $f = 1,000$, the average infeasibility and the optimality gap is zero with the 85\% prediction level. The UC provides a similar average time improvement without infeasibility. When $f = 10,000$, the average time improvement is around 6 and 27 for the instances with $T = 90$ and $T = 120$, respectively, with the prediction level of 85\%. The infeasibility is slightly higher than the instances with $f = 1,000$ since the higher-level predictions increase percent infeasibility for the harder test instances. The UC remedies the infeasibility problem and improves the solution time with a factor of 6 and 28 and with an optimality gap of around 0.7\% and 0.8\% for the instances with $T = 90$ and $T = 120$, respectively. Also, the UC outperforms the MS in time gain for all cases. When looking at the overall averages in Table \ref{t:average}, the LSTM-Opt predictions at the 85\% level reduce the CPLEX solution time by a factor of 9 on average for over 240,000 test instances with an infeasibility below 0.4\% and an optimality gap of less than 0.05\%. The UC provides a similar time gain without any infeasibility and a slightly higher optimality gap of 0.4\% than the 85\% level of prediction.

\renewcommand{\baselinestretch}{1}
\begin{table}[!htb]
\caption{Summary of averages in Tables \ref{t:f10,000t120}, \ref{t:f1,000t90}, \ref{t:f10,000t90}, and \ref{t:f1,000t120}}
\label{t:average}
\centering
\scalebox{0.75}{
\begin{tabular}{ccccccccc}
$f$	&	$T$	&	predicted(\%)	&	timeCPX	&	timeML	&	timeimp	&	timegain(\%)	&	infeasible(\%)	&	optgap(\%)	\\	\hline
1,000	&	90	&	85	&	0.4	&	0.3	&	1	&	18.3	&	0.0	&	0.0	\\	
	&		&	100(MS)	&		&	0.3	&	1	&	3.1	&	0.0	&	0.0	\\	
	&		&	100(UC)	&		&	0.3	&	1	&	25.5	&	0.0	&	0.0	\\	\hline
10,000	&	90	&	85	&	1.8	&	0.3	&	6	&	83.3	&	0.8	&	0.1	\\	
	&		&	100(MS)	&		&	2.1	&	1	&	-12.3	&	0.0	&	0.0	\\	
	&		&	100(UC)	&		&	0.3	&	6	&	83.2	&	0.0	&	0.7	\\	\hline
1,000	&	120	&	85	&	0.4	&	0.3	&	1	&	21.0	&	0.0	&	0.0	\\	
	&		&	100(MS)	&		&	0.4	&	1	&	1.8	&	0.0	&	0.0	\\	
	&		&	100(UC)	&		&	0.3	&	1	&	25.4	&	0.0	&	0.0	\\	\hline
10,000	&	120	&	85	&	8.8	&	0.3	&	27	&	96.2	&	0.7	&	0.1	\\	
	&		&	100(MS)	&		&	1.4	&	6	&	83.9	&	0.0	&	0.4	\\	
	&		&	100(UC)	&		&	0.3	&	28	&	96.4	&	0.0	&	0.8	\\	\hline
	&	Avg.	&	85	&	2.8	&	0.3	&	9	&	54.7	&	0.4	&	0.0	\\	
	&		&	100(MS)	&		&	1.1	&	2	&	19.1	&	0.0	&	0.1	\\	
	&		&	100(UC)	&		&	0.3	&	9	&	57.6	&	0.0	&	0.4	\\	
\end{tabular}}
\end{table}
\renewcommand{\baselinestretch}{}

In summary, the level of predictions used to get the best results varies notably between datasets with different characteristics. This level should be adjusted carefully considering the trade-off between time gain, infeasibility, and optimality gap. Using an appropriate level of predicted variables leads to major reductions in solution time up to an order of magnitude without increasing any infeasibility or optimality gap. It is beneficial to use lower prediction levels for harder instances and higher prediction levels for easier instances, but a prediction level of around 85\% can be a reasonable level for all instances considered in this study. Also, the UC outperforms the MS in terms of providing lower optimality gaps. The UC can also be an alternative to the approach that uses predictions as constraints because it achieves zero infeasibility at the cost of a slightly higher optimality gap. As the $c$, $f$, and $T$ increase, i.e., the instances get harder, and we observe a higher time factor improvement using the LSTM-Opt framework, highlighting the potential of our approach for solving instances with varying sizes and distributions, as discussed in the next section.

\subsection{Results on Generalization}

Here, we present the results on the generalization of our approach to instances with a larger planning horizon $T$. It is not uncommon to have long production planning periods for industries where a daily (even hourly) production planning is necessary, such as large-scale semiconductor manufacturing, and energy production \cite{uzsoy1992review, shrouf2015energy}. Results on different data distributions are presented in Appendix \ref{Predicting Instances with Different Distributions}. We omit the results with the MS approach in favor of UC due to its lack of performance. Generalization is a desired property because it might be beneficial to train the LSTM model in a relatively small horizon to predict instances with a larger planning horizon, saving from the training time. Specifically, the time to train the LSTM model is shorter than the training time for the instance for which the prediction is made due to the smaller number of model parameters. We also compare our framework with two other well-known ML algorithms (logistic regression and random forests) and the state-of-the-art cutting plane algorithms proposed for the CLSP.

\subsubsection{Predicting Instances with Longer Horizons}

Table \ref{gen2} presents the results for predicting datasets with longer planning horizons. The predictions for a larger horizon are generated by concatenating the smaller LSTM predictions obtained by the LSTM model, which has a shorter planning horizon. For example, in the second block of rows in Table \ref{gen2}, the LSTM model with $c = 3$, $f = 1,000$, and $T = 90$ is used to generate predictions for the dataset with the same $c$ and $f$, and $T = 360$. Here, four separate prediction sets, each with 90 periods, are concatenated into a single set of predictions for generating a prediction for the test set with $T = 360$. 

For those instances, predicting 85\% of variables results in a time improvement of 3, with a 0.3\% optimality gap and zero infeasibility in the test set of 20,000 instances. The dataset with $c = 5$, $f = 10,000$, and $T = 180$ is predicted with the LSTM model trained using instances with the same $c$ and $f$ but a half-length planning horizon of $T = 90$, as shown in the third block of rows in Table \ref{gen2}. For those instances, predicting 50\% of variables yields a significant time improvement of 9 and all feasible solutions in the test set at the cost of an optimality gap, which is below 0.5\%. The dataset with $c = 8$, $f = 10,000$, and $T = 480$ constitutes the hardest instances presented in Table \ref{gen2} with the mean solution time over 40 seconds and are predicted using the LSTM model trained with $T = 120$. Here, we observe significant solution time factor improvements over CPLEX using our LSTM-Opt framework. Predictions at the 75\% level reduce the CPLEX solution time by a factor of 25 with no infeasibility and an optimality gap of 1\%.

\renewcommand{\baselinestretch}{1}
\begin{table}[!htb]
\parbox{0.85\textwidth}{\caption{Summary of generalization experiments to test datasets with longer planning horizons\label{gen2}}}
\centering
\scalebox{0.75}{
\begin{tabular}{ccccccccccccc}
\multicolumn{3}{c}{LSTM Train}  					&	\multicolumn{3}{c}{Test Data}					&	       pred  	&	timeCPX	&	timeML	&	timeimp	&	timegain	&	inf	&	optgap	\\	
\cmidrule(lr){1-3}						\cmidrule(lr){4-6}																				
$c$	&	$f$	&	$T$	&	$c$	&	$f$	&	$T$	&	(\%)	&		&		&		&	(\%)	&	(\%)	&	(\%)	\\	\hline
3	&	1,000	&	90	&	3	&	1,000	&	180	&	25	&	0.5	&	0.5	&	1	&	7.6	&	0.0	&	0.1	\\	
        	&	       	&	       	&	               	&	               	&	               	&	50	&		&	0.4	&	1	&	20.5	&	0.0	&	0.1	\\	
        	&	       	&	       	&	               	&	               	&	               	&	75	&		&	0.4	&	1	&	29.1	&	0.0	&	0.2	\\	
        	&	       	&	       	&	               	&	               	&	               	&	85	&		&	0.4	&	1	&	33.3	&	0.0	&	0.2	\\	
        	&	       	&	       	&	               	&	               	&	               	&	90	&		&	0.3	&	2	&	38.2	&	0.0	&	0.2	\\	
        	&	       	&	       	&	               	&	               	&	               	&	95	&		&	0.3	&	2	&	44.2	&	0.1	&	0.2	\\	
        	&	       	&	       	&	               	&	               	&	               	&	100	&		&	0.1	&	6	&	82.4	&	1.0	&	0.4	\\	
        	&	       	&	       	&	               	&	               	&	               	&	100(UC)	&		&	0.3	&	2	&	36.2	&	0.0	&	0.4	\\	\hline
3	&	1,000	&	90	&	3	&	1,000	&	360	&	25	&	1.2	&	1.0	&	1	&	19.7	&	0.0	&	0.1	\\	
        	&	       	&	       	&	               	&	               	&	               	&	50	&		&	0.7	&	2	&	41.6	&	0.0	&	0.2	\\	
        	&	       	&	       	&	               	&	               	&	               	&	75	&		&	0.5	&	2	&	59.6	&	0.0	&	0.2	\\	
        	&	       	&	       	&	               	&	               	&	               	&	85	&		&	0.4	&	3	&	64.2	&	0.0	&	0.3	\\	
        	&	       	&	       	&	               	&	               	&	               	&	90	&		&	0.4	&	3	&	66.4	&	0.0	&	0.3	\\	
        	&	       	&	       	&	               	&	               	&	               	&	95	&		&	0.3	&	3	&	71.4	&	0.2	&	0.3	\\	
        	&	       	&	       	&	               	&	               	&	               	&	100	&		&	0.1	&	12	&	91.9	&	1.3	&	0.5	\\	
        	&	       	&	       	&	               	&	               	&	               	&	100(UC)	&		&	0.4	&	3	&	62.8	&	0.0	&	0.5	\\	\hline
5	&	10,000	&	90	&	5	&	10,000	&	180	&	25	&	19.0	&	6.2	&	3	&	67.2	&	0.0	&	0.3	\\	
        	&	       	&	       	&	               	&	               	&	               	&	50	&		&	2.2	&	9	&	88.4	&	0.0	&	0.5	\\	
        	&	       	&	       	&	               	&	               	&	               	&	75	&		&	0.7	&	28	&	96.4	&	0.1	&	0.5	\\	
        	&	       	&	       	&	               	&	               	&	               	&	85	&		&	0.4	&	46	&	97.8	&	0.3	&	0.6	\\	
        	&	       	&	       	&	               	&	               	&	               	&	90	&		&	0.3	&	59	&	98.3	&	0.9	&	0.7	\\	
        	&	       	&	       	&	               	&	               	&	               	&	95	&		&	0.2	&	78	&	98.7	&	2.8	&	0.9	\\	
        	&	       	&	       	&	               	&	               	&	               	&	100	&		&	0.1	&	166	&	99.4	&	27.5	&	3.5	\\	
        	&	       	&	       	&	               	&	               	&	               	&	100(UC)	&		&	0.4	&	49	&	98.0	&	0.0	&	1.6	\\	\hline
8	&	10,000	&	120	&	8	&	10,000	&	480	&	25	&	42.4	&	11.6	&	4	&	72.7	&	0.0	&	0.6	\\	
        	&	       	&	       	&	               	&	               	&	               	&	50	&		&	4.3	&	10	&	89.9	&	0.0	&	0.9	\\	
        	&	       	&	       	&	               	&	               	&	               	&	75	&		&	1.7	&	25	&	96.1	&	0.0	&	0.9	\\	
        	&	       	&	       	&	               	&	               	&	               	&	85	&		&	0.8	&	55	&	98.2	&	0.0	&	0.9	\\	
        	&	       	&	       	&	               	&	               	&	               	&	90	&		&	0.5	&	86	&	98.8	&	0.1	&	1.0	\\	
        	&	       	&	       	&	               	&	               	&	               	&	95	&		&	0.4	&	112	&	99.1	&	0.4	&	1.0	\\	
        	&	       	&	       	&	               	&	               	&	               	&	100	&		&	0.1	&	364	&	99.7	&	5.2	&	2.8	\\	
        	&	       	&	       	&	               	&	               	&	               	&	100(UC)	&		&	0.6	&	71	&	98.6	&	0.0	&	1.5	\\	\hline
\multicolumn{13}{l}{$^{*}$\footnotesize{Experiments include 20,000 test instances.}}

\end{tabular}}
\end{table}
\renewcommand{\baselinestretch}{}

Table \ref{gen3}  presents the results for predicting datasets with significantly longer planning horizons; therefore, the test instances are much harder than the training instances. The predictions are generated using the model trained with $c = 8$, $f = 10,000$, and $T = 120$. The test sets for all three datasets consist of 10 instances, instead of 20,000 as previously presented, due to computational complexity and long solution times. For the first dataset with $T = 600$, problems are solved 70 times faster than the default CPLEX using predictions at the 50\% level without any infeasibility and with an optimality gap below 1\%. The mean CPLEX solution time for the next dataset with $T = 720$ is more than 8 CPU hours. Here, the solution time of 8 hours is reduced to under 1 minute, with the predictions used at the 25\% level without any infeasibility and with an optimality gap below 1\%. Predictions used at 75\% reduce the solution time by more than four orders of magnitude from more than 8 CPU hours to 2.5 CPU seconds without infeasibility and with an optimality gap below 2\%. The last test dataset with $c = 8$, $f = 10,000$, and $T = 960$ constitutes the hardest instances presented with a mean solution time of over 70 hours using CPLEX at its default settings. For those instances, predictions at the 25\% level reduce the solution time of 70 CPU hours to only 79 CPU seconds with an optimality gap of 0.8\% and without any infeasibility. Predictions at the 50\% level reduce the solution time by more than a factor of 16,000, with an optimality gap of 1.2\% and zero infeasibility. 

Generating 100,000 instances for training data with $c = 8$, $f = 10,000$, and $T = 120$ takes 140,170 seconds whereas the LSTM training time takes 52,724 seconds. In this specific example, it can be concluded that generating training data, training the LSTM model, and resolving with predictions for a single instance takes 16 hours less than solving the instance with CPLEX. For such hard problems, our LSTM-Opt framework achieves a significant time reduction even in the case where just a single instance must be solved. The results discussed above highlight that our approach could be generalizable to predict larger instances with substantial benefits in reducing the solution time of those hard CLSPs with the cost of a small optimality gap.

\renewcommand{\baselinestretch}{1}
\begin{table}[!htb]
\parbox{0.85\textwidth}{\caption{Summary of generalization experiments to test datasets with longer planning horizons contd.$^{*}$\label{gen3}}}
\centering
\scalebox{0.75}{
\begin{tabular}{ccccccccccccc}
\multicolumn{3}{c}{LSTM Train}  					&	\multicolumn{3}{c}{Test Data}					&	       pred  	&	timeCPX	&	timeML	&	timeimp	&	timegain	&	inf	&	optgap	\\	
\cmidrule(lr){1-3}						\cmidrule(lr){4-6}																				
$c$	&	$f$	&	$T$	&	$c$	&	$f$	&	$T$	&	(\%)	&		&		&		&	(\%)	&	(\%)	&	(\%)	\\	\hline
8	&	10,000	&	120	&	8	&	10,000	&	600	&	25	&	 409 	&	31.1	&	13	&	92.4	&	0.0	&	0.6	\\	
        	&	       	&	       	&	               	&	               	&	               	&	50	&		&	5.8	&	70	&	98.6	&	0.0	&	0.9	\\	
        	&	       	&	       	&	               	&	               	&	               	&	75	&		&	2.5	&	161	&	99.4	&	0.0	&	1.2	\\	
        	&	       	&	       	&	               	&	               	&	               	&	85	&		&	1.2	&	343	&	99.7	&	0.0	&	1.5	\\	
        	&	       	&	       	&	               	&	               	&	               	&	90	&		&	0.9	&	476	&	99.8	&	0.0	&	1.7	\\	
        	&	       	&	       	&	               	&	               	&	               	&	95	&		&	0.6	&	712	&	99.9	&	0.0	&	2.0	\\	
        	&	       	&	       	&	               	&	               	&	               	&	100	&		&	0.2	&	1,990	&	99.9	&	0.0	&	3.0	\\	
        	&	       	&	       	&	               	&	               	&	               	&	100(UC)	&		&	6.0	&	68	&	98.5	&	0.0	&	1.6	\\	\hline
8	&	10,000	&	120	&	8	&	10,000	&	720	&	25	&	 30,038 	&	54.5	&	552	&	99.8	&	0.0	&	0.9	\\	
        	&	       	&	       	&	               	&	               	&	               	&	50	&		&	7.2	&	4,168	&	100.0	&	0.0	&	1.2	\\	
        	&	       	&	       	&	               	&	               	&	               	&	75	&		&	2.5	&	12,014	&	100.0	&	0.0	&	1.7	\\	
        	&	       	&	       	&	               	&	               	&	               	&	85	&		&	1.0	&	28,888	&	100.0	&	0.0	&	2.1	\\	
        	&	       	&	       	&	               	&	               	&	               	&	90	&		&	0.8	&	38,788	&	100.0	&	0.0	&	2.5	\\	
        	&	       	&	       	&	               	&	               	&	               	&	95	&		&	0.6	&	51,267	&	100.0	&	0.0	&	3.0	\\	
        	&	       	&	       	&	               	&	               	&	               	&	100	&		&	0.2	&	164,035	&	100.0	&	10.0	&	6.4	\\	
        	&	       	&	       	&	               	&	               	&	               	&	100(UC)	&		&	9.7	&	3,150	&	100.0	&	1.4	&	2.5	\\	\hline
8	&	10,000	&	120	&	8	&	10,000	&	960	&	25	&	 252,186 	&	78.6	&	3,208	&	100.0	&	0.0	&	0.8	\\	
        	&	       	&	       	&	               	&	               	&	               	&	50	&		&	15.2	&	16,543	&	100.0	&	0.0	&	1.2	\\	
        	&	       	&	       	&	               	&	               	&	               	&	75	&		&	3.1	&	80,922	&	100.0	&	0.0	&	1.6	\\	
        	&	       	&	       	&	               	&	               	&	               	&	85	&		&	1.3	&	199,593	&	100.0	&	0.0	&	2.0	\\	
        	&	       	&	       	&	               	&	               	&	               	&	90	&		&	0.8	&	310,612	&	100.0	&	0.0	&	2.3	\\	
        	&	       	&	       	&	               	&	               	&	               	&	95	&		&	0.6	&	411,262	&	100.0	&	0.0	&	2.7	\\	
        	&	       	&	       	&	               	&	               	&	               	&	100	&		&	0.2	&	1,245,361	&	100.0	&	0.0	&	5.6	\\	
        	&	       	&	       	&	               	&	               	&	               	&	100(UC)	&		&	14.3	&	17,678	&	100.0	&	0.0	&	2.3	\\	\hline
\multicolumn{13}{l}{$^{*}$\footnotesize{Experiments only include ten test instances due to long solution times.}}

\end{tabular}}
\end{table}
\renewcommand{\baselinestretch}{}

\subsection{Comparison with other ML and exact algorithms}

Table \ref{gen4} presents the computational comparison of our LSTM-Opt framework with two other machine learning approaches that perform a binary classification task and shows that their prediction quality is not comparable to LSTM-Opt. Additionally, the comparison with two other exact approaches is presented in Table \ref{gen4extra} to show that the LSTM-Opt framework could produce good-quality solutions in much less time compared to those exact approaches. The machine learning and exact approaches used to compare with the LSTM-Opt are defined as follows. 

ML Approaches:

\begin{itemize}
\item Logistic Regression (LR): An extension of linear regression, is more interpretable than the tree-based ensemble methods such as random forest at the cost of accuracy.
\item Random Forest (RF): One of the best algorithms for classification tasks \citep{fernandez2014we} in terms of classification accuracy at the cost of reduced
interpretability.

\end{itemize}

Exact Approaches:

\begin{itemize}

\item CPLEX (CPX): Direct solution of the CLSP formulation \eqref{objective1-ex:1}-\eqref{integrality1-e:1} with default CPLEX.

\item Dynamic programming-based inequalities (DPineq) of \citet{hartman2010dynamic}: We used the weaklu strategy to create a tighter CLSP polyhedron. The generated inequalities are added to the formulation \eqref{clsp:1}, and the proposed algorithm is shown to outpace the dynamic programming algorithm by \citet{hartman2010dynamic} for some cases. In the experiments, we generate cuts for the first 100 periods with $c = 3$, for the first 75 periods with $c = 5$, and for the first 50 periods with $c = 8$ for instances with $T = 360$ to combat the growing DP-based inequality generation time with increasing $c$.

\item The ($\ell$, S) inequalities (LSineq) of \citet{barany1984strong}: Implemented with a separation algorithm since the number of ($\ell$, S) inequalities grows exponentially. The separation algorithm is iterated five times which is inclined to give the best computational achievements \citep{buyuktahtakin2018partial}.

\item Dynamic programming (DP) solution approach for CLSP \citep{hartman2010dynamic,florian1980deterministic}: Results are omitted from the tables due to lack of performance. For example, while the mean solution times of $c = 3,5,8$ instances with CPLEX were 22.4, 3.3, and 1.3 seconds, respectively, the dynamic programming solution times were 610.7, 1054.9, and 1938.8 seconds for the same first 20 instances presented in Table \ref{t:f10,000t120}. Additionally, the dynamic programming approach has a complexity of $\mathcal{O}(TD_{T}^2)$ where $D_{T} = \sum_{t=1}^{T} d_{t}$. We do not further include the dynamic programming solution to compare with the LSTM-Opt framework since CPLEX is superior for the considered instances.

\end{itemize}

Table \ref{gen4}  presents a set of instances that are tested for the comparison of LSTM-Opt, LR, and RF. Here, we have utilized a different structure to generate our test instances. For the instances with $c = 3,5$, a solution time limit of 86,400 CPU seconds (24 CPU hours) is set for CPLEX to restrict the solution time. The metrics, including time improvement, time gain, and optimality gap, are calculated based on the best solution found by CPLEX within the solution time limit. The $IGap=100 \times (objCPX-objLP)/objCPX$, where $objCPX$ is the objective function value of the best feasible solution to the original problem and $objLP$ is the objective function to its linear programming relaxation, is 7.5\%, 16.1\%, and 27.7\% for the instances with $c=3,5,$ and $8$, respectively. CPLEX reports an MIP optimality gap of 0.64\%, 0.06\%, and 0.00\% on average for test instances with $c = 3,5,$ and $8$, respectively, with a one-day time limit. For the same instances with $c = 3,5$, our preliminary results revealed that the test problems were still computationally very expensive to solve without a time limit with the 25\% prediction level. Therefore the results with the 25\% prediction level are omitted from the results in this section. 

In Table \ref{gen4}, for the $c = 3$ instances, predictions at the 50\% level improve the solution time by more than a factor of 7,500 by reducing the limited average solution time from one day to only 12 seconds, without any infeasibility and with an optimality gap of 0.8\%. Predictions of more than 50\% lead to some infeasibility in the test set, while the predictions at the 100\% level lead to all infeasible predictions, and thus the corresponding results are presented in a “-” in the first-row block of results in Table \ref{gen4}. For the same instances solved at the 50\% prediction level, LR and RF have caused an infeasibility of 70\% and 50\%, respectively, since neither considers sequential dependency like the LSTM networks. The optimality gaps of the feasible instances were significantly higher than the 0.8\% of the LSTM-Opt framework at 1.9\% and 2.3\%, respectively, for both LR and RF. Also, the time improvements are not as big as the ones of the LSTM-Opt framework. For the instances with $c = 5$, predictions using LSTM-Opt at the 50\% level decrease the limited solution time from 1-day to 20 CPU seconds and reduce the solution time by more than a factor of 4,000, including the prediction generation time without any infeasibility and with an optimality gap of 0.9\%. The 85\% level with LSTM-Opt reduces the solution time by five orders of magnitude without infeasibility and with an optimality gap of 2.3\%. For the same prediction level, both the LR ad RF causes all infeasible predictions. The datasets with $c = 8$ constitute significantly faster to solve compared to instances with $c = 3,5$, and the results resemble the structure with easier instances. The LR and RF cause high infeasibility in all prediction levels compared to the LSTM-Opt framework. Table \ref{gen4} shows that a significant solution time reduction of around four to five orders of magnitude is achieved by LSTM-Opt without any infeasibility depending on the desired optimality gap. We anticipate that the solution time gains would have further increased if the original solution time was not limited to 24 hours. Additionally, both the LR and RF cause higher infeasibility, a higher optimality gap, and a lower time improvement.

Table \ref{gen4extra} presents a comparison of LSTM-Opt at the 50\% prediction level with two exact approaches, namely ($\ell$, S) valid inequalities (LSineq) and DP-based cutting planes (DPineq). The solution time is limited by 1 hour, including the inequality generation time for both approaches. Here, the time improvement metric has been calculated with respect to the set solution time of 1-hour. The optimality gap metric has been calculated with respect to the best integer solution found by the CPLEX within a 1-day solution time limit. For the $c=3$ instances, the LSTM-Opt framework achieves a 1.6\% optimality gap. The objective function value is reduced below the 1-day limited CPLEX solution value with a 1-hour time limit in both formulations with DP-based and ($\ell$, S) inequalities, resulting in a negative optimality gap of -0.01\%. Therefore, both types of inequalities are effective at reducing the optimality gap, but they can not solve the CLSP very fast though they speed up the solution for harder instances of $c = 3,5$. Even though the LSTM-Opt framework has an optimality gap of 0.8\%, the solution was found 322 times faster than CPLEX, showing that the LSTM-Opt can solve those problems very fast. The results with $c = 5$ show a similar pattern, and the LSTM-Opt framework has an optimality gap of 0.9\% on top of the CPLEX gap. The results with $c = 8$ show that LSTM-Opt framework can achieve a time improvement of 3 while inequality-based methods increase the solution time for easier instances.

Figures \ref{figcompall1} and \ref{figcompall1} present a summary of the results for the instances presented in Tables \ref{gen4} and \ref{gen4extra}, for $c=3$ and $5$, respectively. Both DPineq and LSineq improve over the CPLEX gap in Figure \ref{figcompall1} with $c = 3$ instances. LSTM achieves a solution with a slightly larger optimality gap much faster than those exact inequality methods and also is faster than LR and RF with a lower gap. The latter two algorithms result in high infeasibility of 70\% and 50\%, respectively, while LSTM-Opt achieves all-feasible predictions. The results show similarity for the $c = 5$ instances in Figure \ref{figcompall2}, with the exception that both exact methods do not improve over the CPLEX gap or solution time. Even though LR and RF have a higher time improvement than LSTM-Opt, both lead to high and unpractical infeasibility rates of 60\% and 40\%, respectively.

\renewcommand{\baselinestretch}{1.0}
\begin{landscape}
\begin{table}[!htb]
\parbox{1.4\textwidth}{\caption{Computational results for comparing LSTM-Opt with other ML algorithms$^{*}$\label{gen4}}}
\centering
\scalebox{0.78}{
\begin{tabular}{cccccccccccccccccc}

	&		&	Train	&	Test	&	       pred  	&	timeCPX	&	time	&	timeimp	&	inf(\%)	&	optgap(\%)	&	time	&	timeimp	&	inf(\%)	&	optgap(\%)	&	time	&	timeimp	&	inf(\%)	&	optgap(\%)	\\	
				\cmidrule(lr){3-3}		\cmidrule(lr){4-4}						\cmidrule(lr){7-10}								\cmidrule(lr){11-14}								\cmidrule(lr){15-18}								
$c$	&	$f$	&	$T$	&	$T$	&	(\%)	&		&	\multicolumn{4}{c}{LSTM-Opt}							&	\multicolumn{4}{c}{LR}							&	\multicolumn{4}{c}{RF}							\\	\hline
3	&	10,000	&	90	&	360	&	50	&	 86,415 	&	11.3	&	7,675	&	0.0	&	0.8	&	42.7	&	 2,024 	&	70.0	&	1.9	&	76.6	&	 1,129 	&	50.0	&	2.3	\\	
        	&	       	&	       	&	               	&	75	&		&	1.4	&	63,189	&	10.0	&	2.0	&	0.6	&	 146,459 	&	90.0	&	17.3	&	-	&	 - 	&	100.0	&	-	\\	
        	&	       	&	       	&	               	&	85	&		&	0.6	&	139,042	&	20.0	&	2.9	&	-	&	 - 	&	100.0	&	-	&	-	&	 - 	&	100.0	&	-	\\	
        	&	       	&	       	&	               	&	90	&		&	0.5	&	159,647	&	30.0	&	3.6	&	-	&	 - 	&	100.0	&	-	&	-	&	 - 	&	100.0	&	-	\\	
        	&	       	&	       	&	               	&	95	&		&	0.4	&	193,403	&	50.0	&	4.5	&	-	&	 - 	&	100.0	&	-	&	-	&	 - 	&	100.0	&	-	\\	
        	&	       	&	       	&	               	&	100	&		&	-	&	-	&	100.0	&	-	&	-	&	 - 	&	100.0	&	-	&	-	&	 - 	&	100.0	&	-	\\	\hline
5	&	10,000	&	90	&	360	&	50	&	 86,763 	&	20.4	&	4,253	&	0.0	&	0.9	&	8.2	&	 10,552 	&	60.0	&	1.4	&	13.4	&	 6,491 	&	40.0	&	1.9	\\	
        	&	       	&	       	&	               	&	75	&		&	2.5	&	34,371	&	0.0	&	1.6	&	0.7	&	 124,883 	&	80.0	&	25.8	&	0.6	&	 145,415 	&	90.0	&	24.2	\\	
        	&	       	&	       	&	               	&	85	&		&	0.8	&	102,727	&	0.0	&	2.3	&	-	&	 - 	&	100.0	&	-	&	-	&	 - 	&	100.0	&	-	\\	
        	&	       	&	       	&	               	&	90	&		&	0.7	&	118,189	&	0.0	&	2.7	&	-	&	 - 	&	100.0	&	-	&	-	&	 - 	&	100.0	&	-	\\	
        	&	       	&	       	&	               	&	95	&		&	0.5	&	190,865	&	10.0	&	3.3	&	-	&	 - 	&	100.0	&	-	&	-	&	 - 	&	100.0	&	-	\\	
        	&	       	&	       	&	               	&	100	&		&	0.2	&	416,680	&	60.0	&	14.2	&	-	&	 - 	&	100.0	&	-	&	-	&	 - 	&	100.0	&	-	\\	\hline
8	&	10,000	&	90	&	360	&	50	&	7.5	&	2.6	&	3	&	0.0	&	1.6	&	3.2	&	 2 	&	40.0	&	0.5	&	3.1	&	 2 	&	30.0	&	1.0	\\	
        	&	       	&	       	&	               	&	75	&		&	1.0	&	8	&	10.0	&	1.9	&	0.7	&	 8 	&	80.0	&	14.3	&	0.7	&	 8 	&	90.0	&	22.5	\\	
        	&	       	&	       	&	               	&	85	&		&	0.7	&	11	&	10.0	&	2.2	&	0.4	&	 14 	&	90.0	&	111.1	&	-	&	 - 	&	100.0	&	-	\\	
        	&	       	&	       	&	               	&	90	&		&	0.6	&	13	&	20.0	&	2.4	&	-	&	 - 	&	100.0	&	-	&	-	&	 - 	&	100.0	&	-	\\	
        	&	       	&	       	&	               	&	95	&		&	0.5	&	16	&	20.0	&	2.8	&	-	&	 - 	&	100.0	&	-	&	-	&	 - 	&	100.0	&	-	\\	
        	&	       	&	       	&	               	&	100	&		&	0.2	&	38	&	20.0	&	3.8	&	-	&	 - 	&	100.0	&	-	&	-	&	 - 	&	100.0	&	-	\\	\hline

\multicolumn{13}{l}{$^{*}$\footnotesize{Experiments only include ten test instances due to long solution times.}}

\end{tabular}}
\end{table}

\begin{table}[!htb]
\parbox{1.4\textwidth}{\caption{Computational results for comparing LSTM-Opt with different exact methods$^{*}$\label{gen4extra}}}
\centering
\scalebox{0.8}{
\begin{tabular}{ccccccccccccccccc}
	&		&	Train	&	Test	&	       pred  	&	inf	&	\multicolumn{4}{c}{$time$}							&	\multicolumn{3}{c}{$timeimp$}					&	\multicolumn{4}{c}{$optgap(\%)$}							\\	
				\cmidrule(lr){3-3}		\cmidrule(lr){4-4}						\cmidrule(lr){7-10}								\cmidrule(lr){11-13}						\cmidrule(lr){14-17}								
$c$	&	$f$	&	$T$	&	$T$	&		&	(\%)	&	CPX	&	Dpineq	&	LSineq	&	LSTM-Opt	&	Dpineq	&	LSineq	&	LSTM-Opt	&	CPX	&	Dpineq	&	LSineq	&	LSTM-Opt	\\	\hline
3	&	10,000	&	90	&	360	&	50	&	0.0	&	 3,602 	&	 3,609 	&	 3,602 	&	11.2	&	1	&	1	&	322	&	0.03	&	-0.1	&	-0.1	&	0.8	\\	
5	&	10,000	&	90	&	360	&	50	&	0.0	&	 3,602 	&	 3,421 	&	 3,600 	&	20.3	&	1	&	1	&	177	&	0.05	&	0.0	&	0.0	&	0.9	\\	
8	&	10,000	&	90	&	360	&	50	&	0.0	&	 7.5 	&	 720 	&	 146 	&	2.5	&	0	&	0	&	3	&	0.00	&	0.0	&	0.0	&	1.6	\\	\hline
\multicolumn{16}{l}{$^{*}$\footnotesize{Experiments only include ten test instances and limited with 1-hour time limit due to long solution times.}}

\end{tabular}}
\end{table}

\end{landscape}
\renewcommand{\baselinestretch}{}

\begin{figure}[!htbp]
\centering
\caption{Comparison of exact and ML algorithms}\label{figcompall}
\addtocounter{figure}{-1}
\subfloat[Instances with $c=3$, $f = 10,000$, and $T = 360$]{%
  \includegraphics[width=0.49\textwidth]{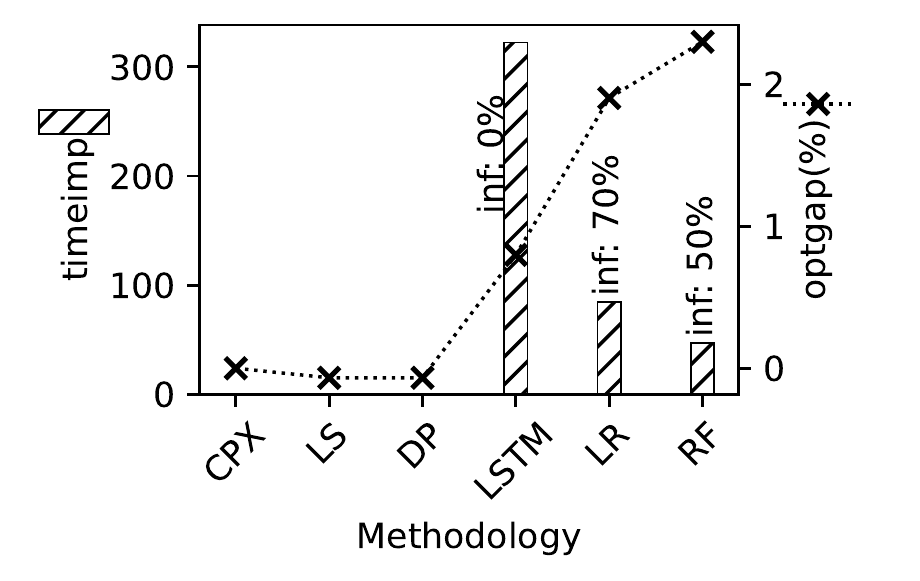}%
  \label{figcompall1}
}%
\subfloat[Instances with $c=5$, $f = 10,000$, and $T = 360$]{%
  \includegraphics[width=0.49\textwidth]{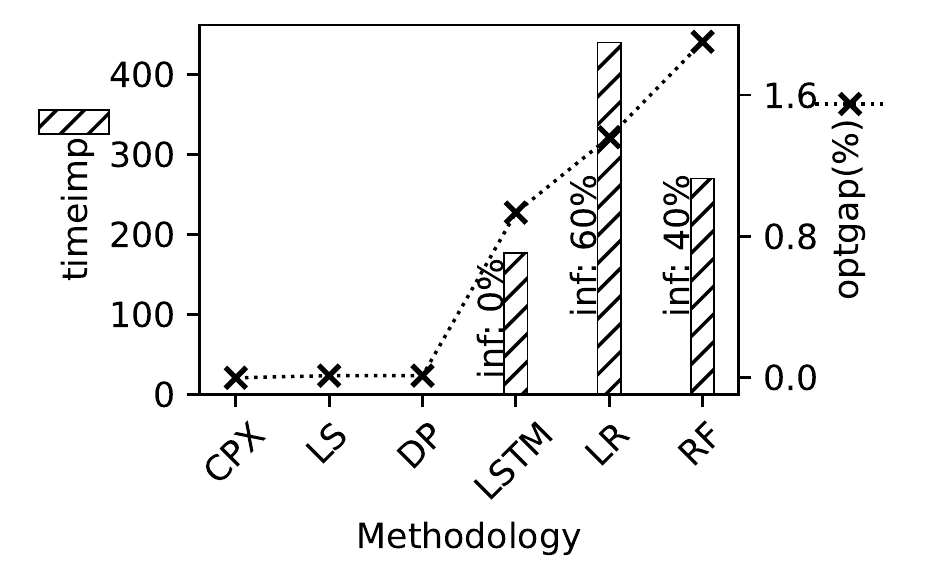}%
  \label{figcompall2}
}%

\stepcounter{figure}

\end{figure}

\subsection{Summary of Results}

The results presented on generalization experiments show that a network trained on a smaller planning horizon can be used to successfully predict the optimal solutions of the instances with larger horizons without any additional training. The solution time can be reduced up to six orders of magnitude without increasing the optimality gap or infeasibility much, especially in harder problems. Also, LSTM-Opt can capture sequential dependencies while LR and RF can not. Classical exact approaches can not produce very fast solutions like the LSTM-Opt.


Figures \ref{figgen22}-\ref{figgen42} present the results for datasets for longer planning horizons. The results for $T = 360$ in Figure \ref{figgen22} are similar to dataset where LSTM model is trained with $c = 3$, $f = 1,000$, and $T = 90$, as shown in Figure \ref{figT90f1000}. For the dataset with $c = 8$, $f = 10,000$, and $T = 720$ in Figure \ref{figgen26}, using predictions at the 50\% level reduces the solution time from more than 8 hours to under 8 seconds without any infeasibility and with an optimality gap of 1.2\%. Figure \ref{figgen27} constitutes the instances with the longest solution times. The instances with $c = 8$, $f = 10,000$, and $T = 960$ have a mean solution time over 70 hours. Predictions at the 25\% and 50\% levels reduce the solution time of those instances by more than a factor of 3,000 and 16,000 with an optimality gap of 0.8\% and 1.2\%, without any infeasibility in the test set, respectively.

Overall averages in Table \ref{gen3} show that the solution time can be decreased with a factor of more than 9,000 with an infeasibility in the test set of only 0.5\% and an optimality gap of approximately 2.1\%. The UC reduces the solution time by more than a factor of 90,000 on average without infeasibility in the test set and an average optimality gap of around 3.5\%. Overall, predictions at the levels between 25\% and 85\% provide significant time improvements with less than a 1\% optimality gap and without any infeasibility in the test set. Specifically, predictions at the 85\% level can balance a high solution time factor improvement with infeasibility and optimality gap at reasonable levels when predicting for longer periods using the LSTM model trained with instances of shorter and thus easier instances.

\begin{figure}[!htbp]
\centering
\caption{Summary of Generalization Experiments}\label{figresultsgen}
\addtocounter{figure}{-1}
\subfloat[LSTM trained with $c=3$, $f = 1,000$, $T = 90$ predicts $c=3$, $f = 1,000$, $T = 360$\label{figgen22}]{%
  \includegraphics[width=0.44\textwidth]{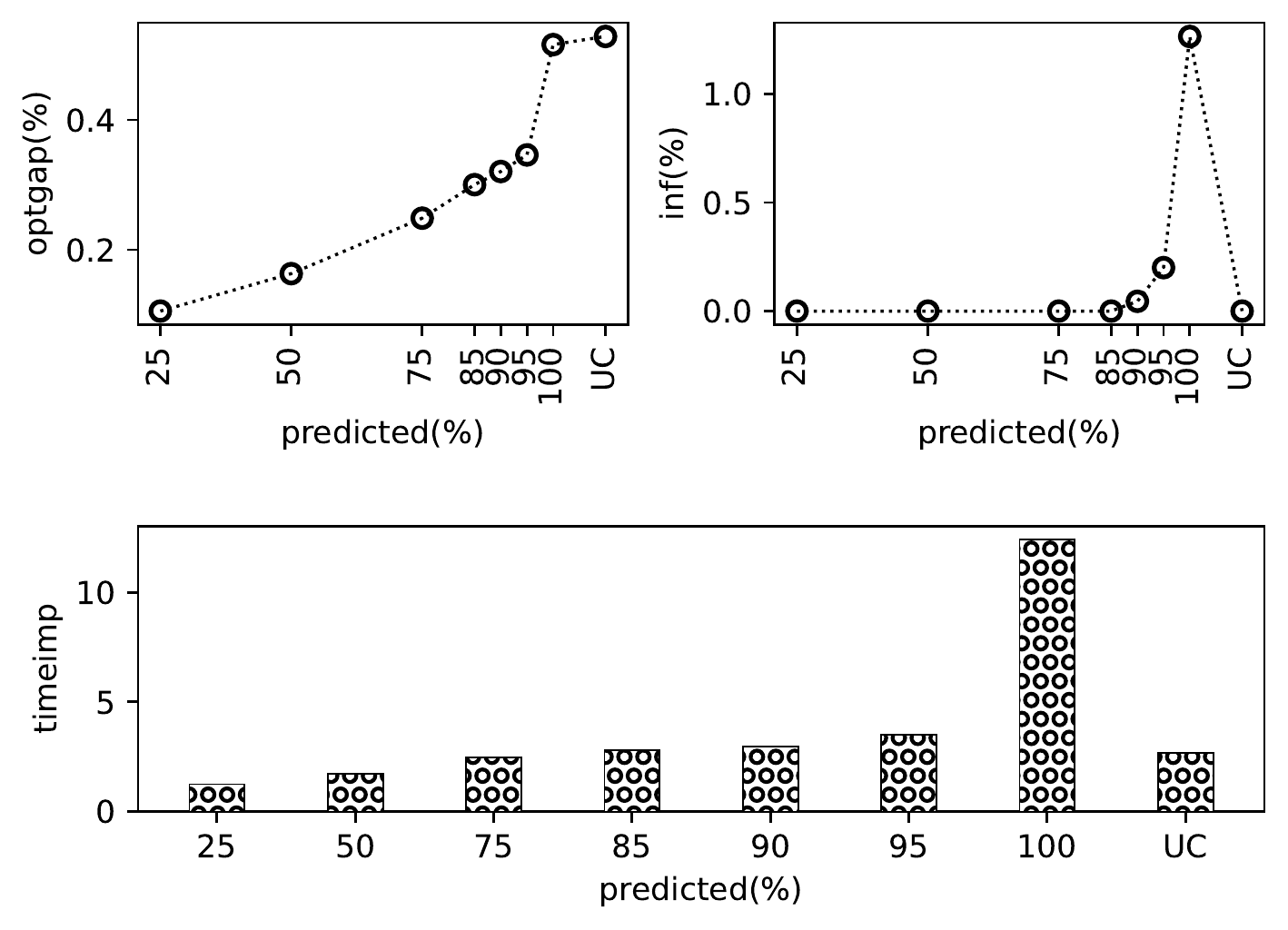}%
}%
\subfloat[LSTM trained with $c=5$, $f = 10,000$, $T = 90$ predicts $c=5$, $f = 10,000$, $T = 180$\label{figgen23}]{%
  \includegraphics[width=0.44\textwidth]{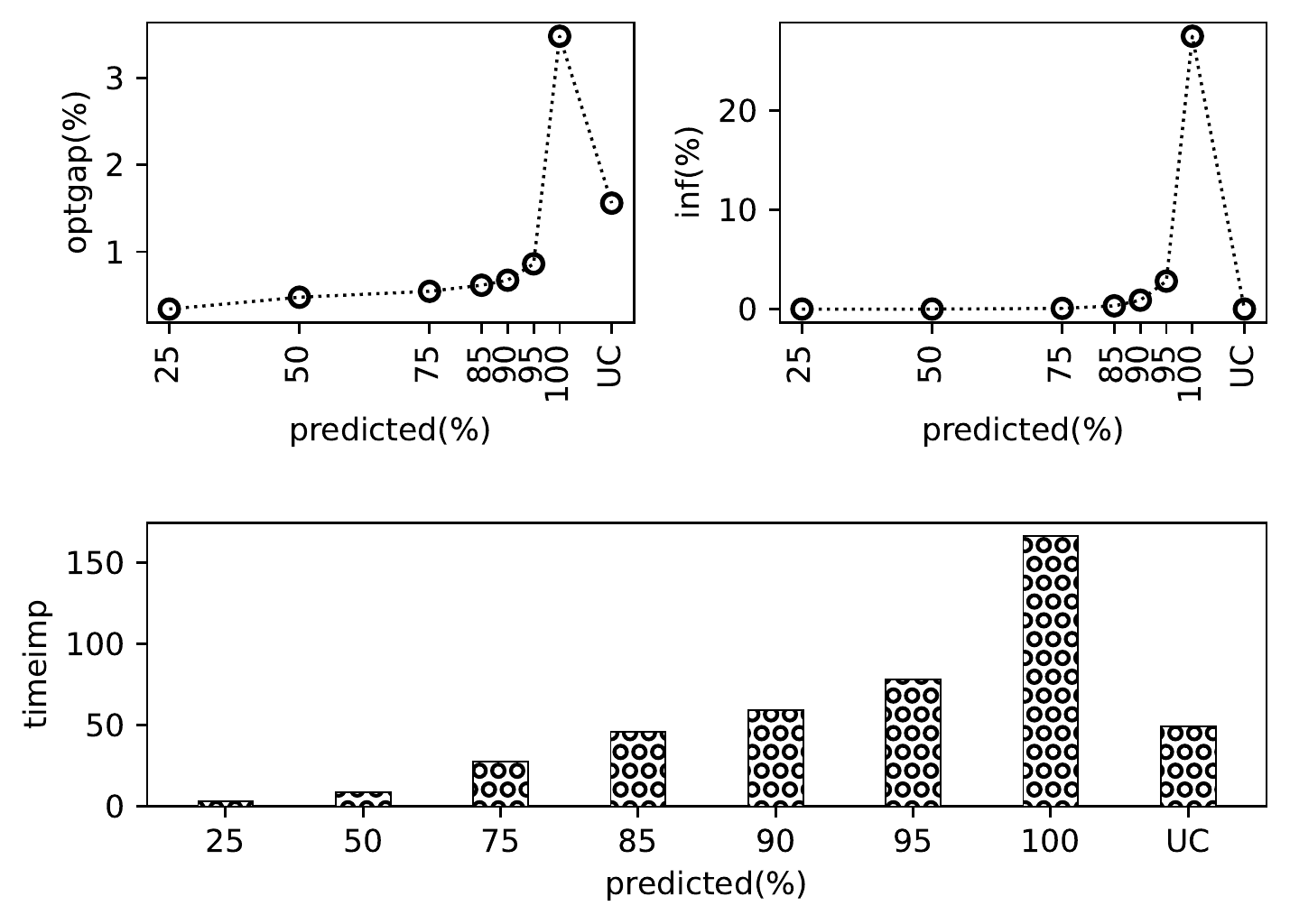}%
}%

\subfloat[LSTM trained with $c=8$, $f = 10,000$, $T = 120$ predicts $c=8$, $f = 10,000$, $T = 720$\label{figgen26}]{%
  \includegraphics[width=0.44\textwidth]{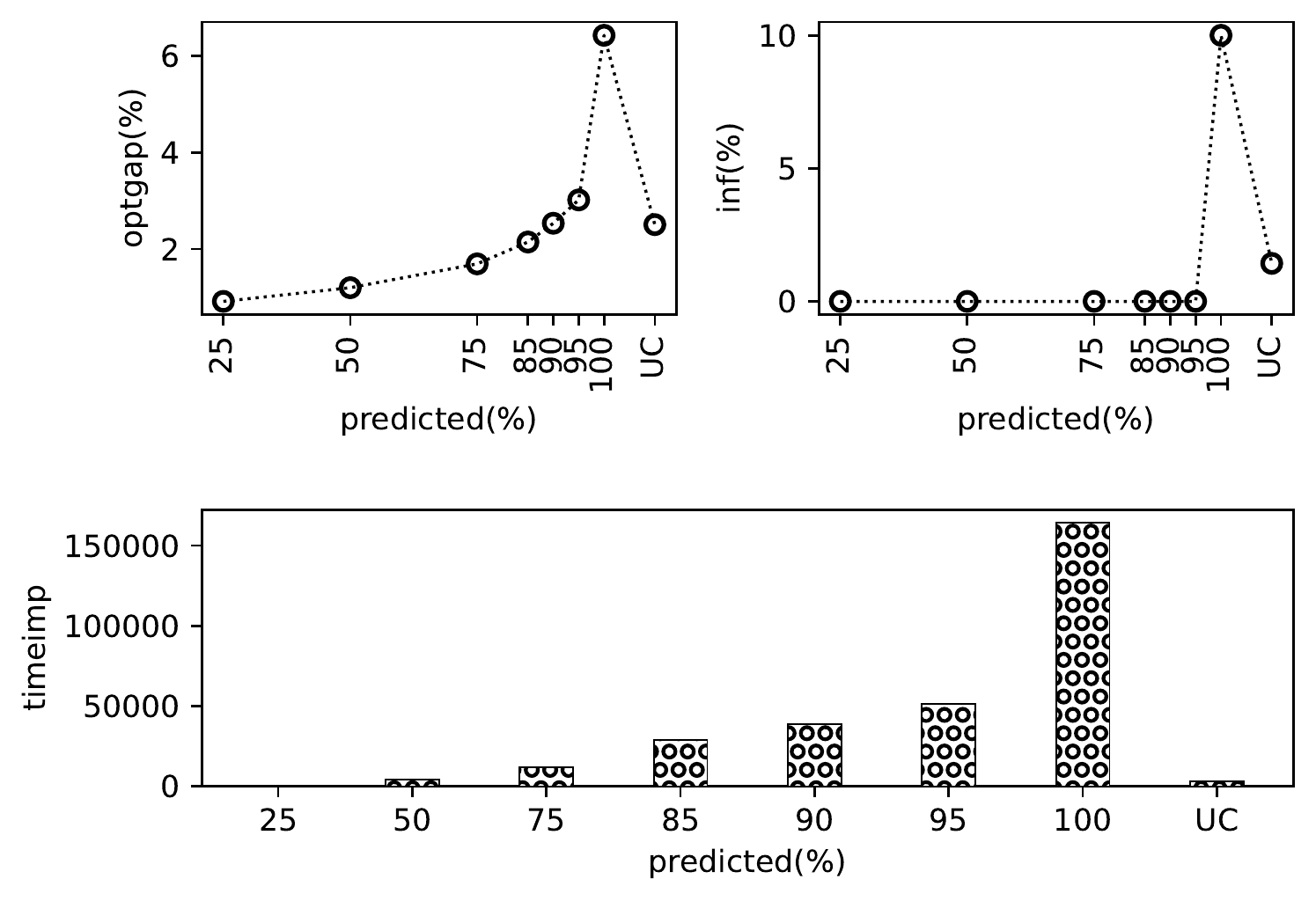}%
}%
\subfloat[LSTM trained with $c=8$, $f = 10,000$, $T = 120$ predicts $c=8$, $f = 10,000$, $T = 960$\label{figgen27}]{%
  \includegraphics[width=0.44\textwidth]{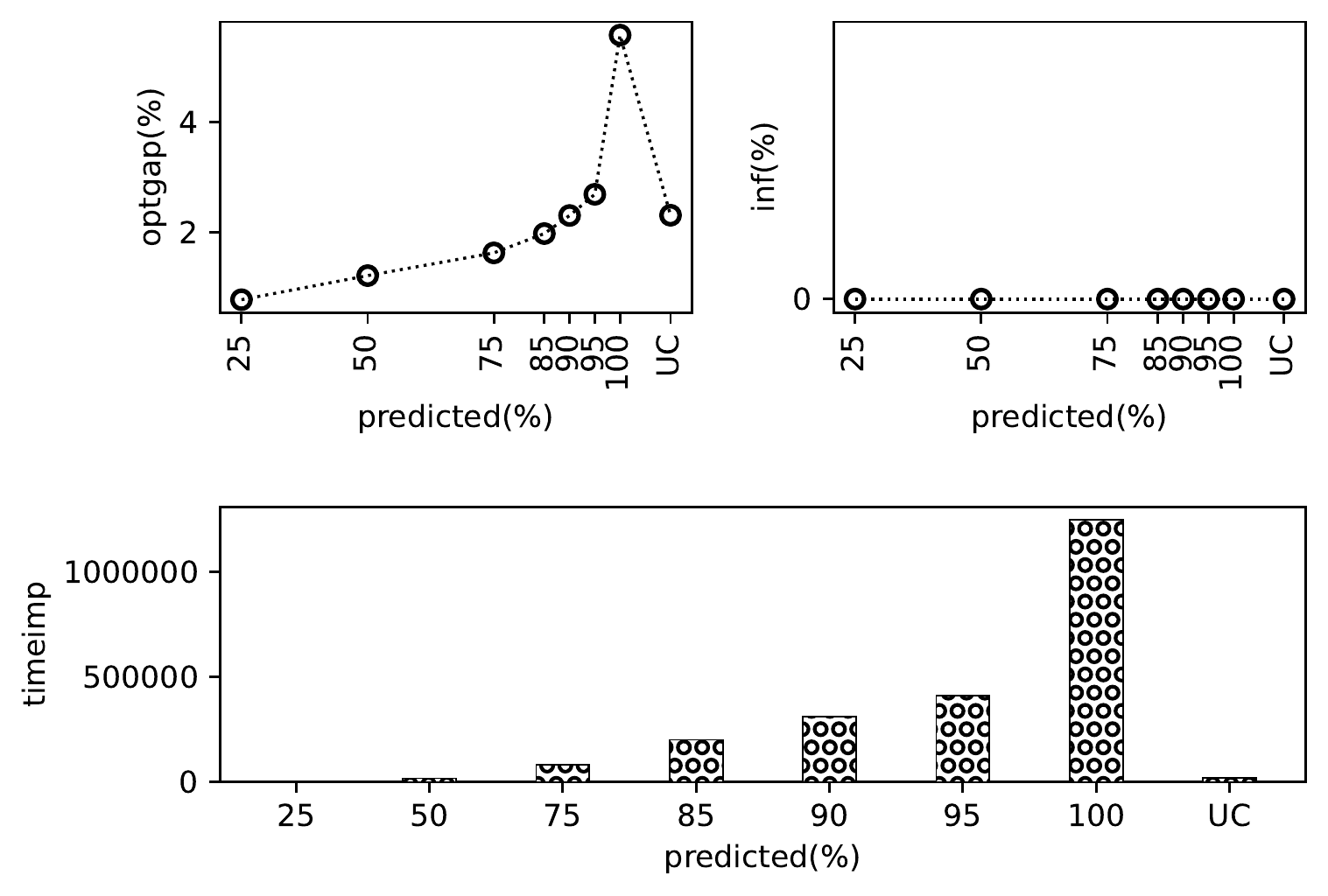}%
}%

\subfloat[LSTM trained with $c=3$, $f = 10,000$, $T = 90$ predicts $c=3$, $f = 10,000$, $T = 360$\label{figgen41}]{%
  \includegraphics[width=0.44\textwidth]{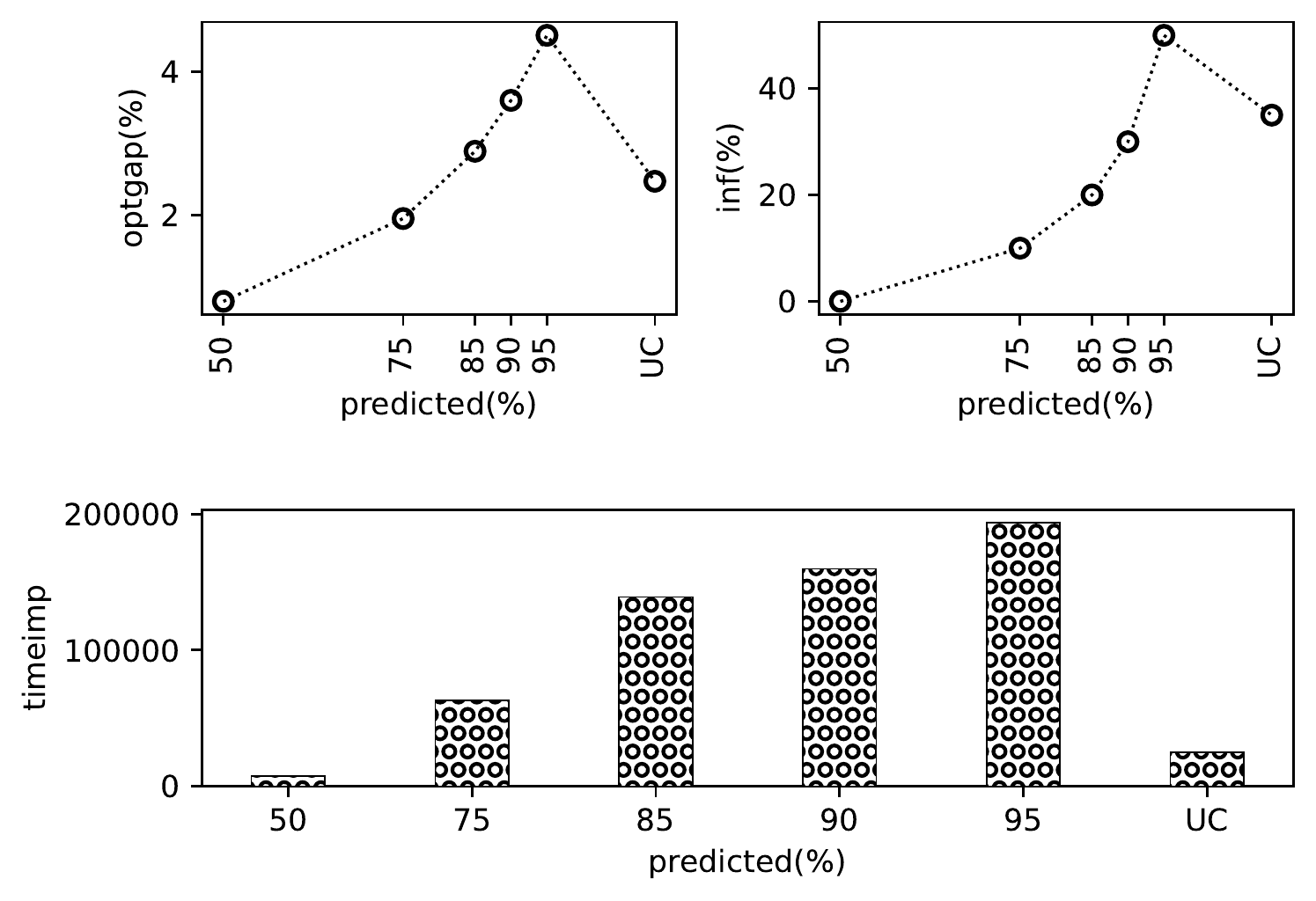}%
}%
\subfloat[LSTM trained with $c=5$, $f = 10,000$, $T = 90$ predicts $c=5$, $f = 10,000$, $T = 360$\label{figgen42}]{%
  \includegraphics[width=0.44\textwidth]{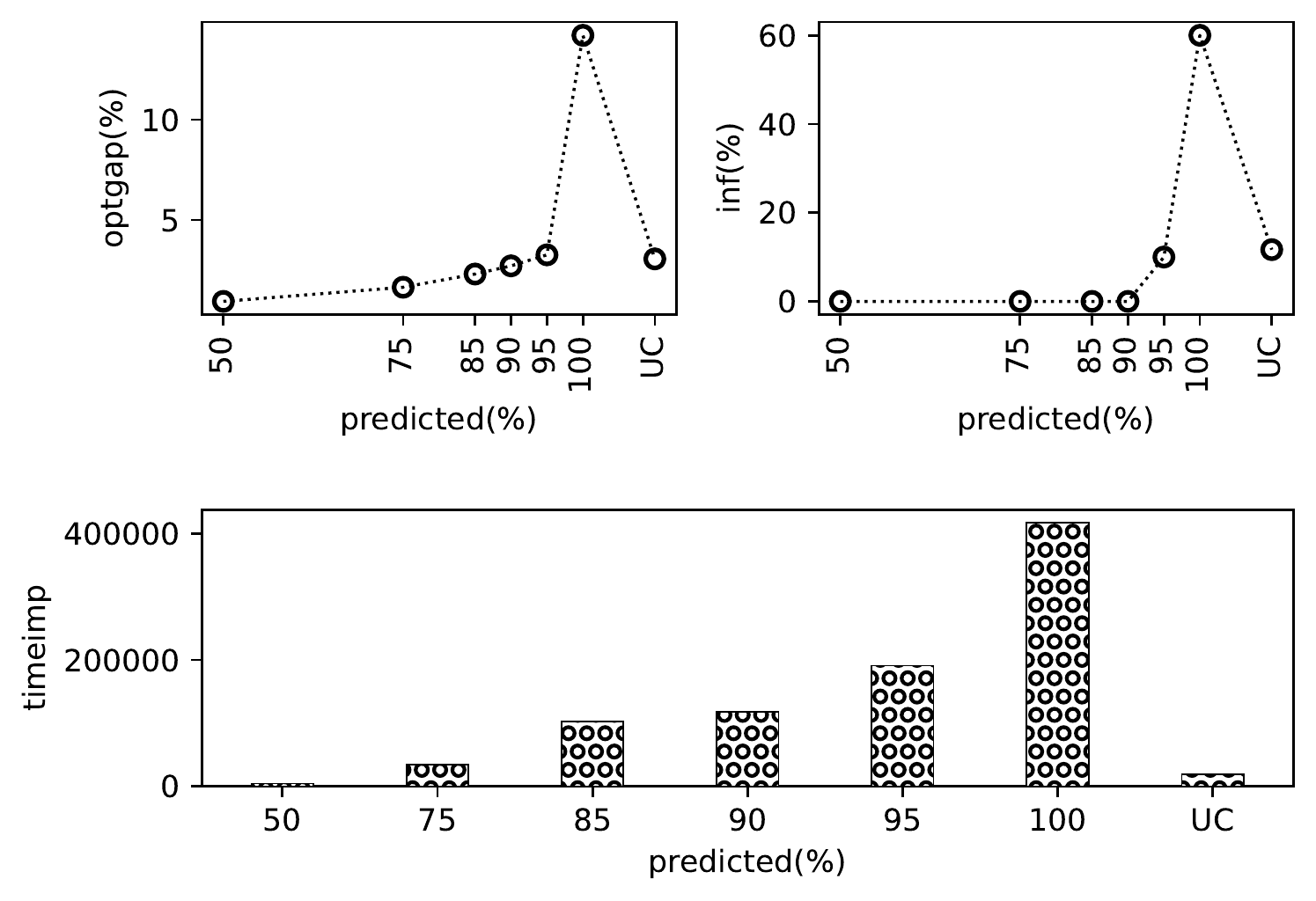}%
}%
\stepcounter{figure}

\end{figure}

\section{Conclusions and Future Work}
\label{Conclusions and Future Work}		

In this study, we present a new LSTM-Opt framework to predict the optimal solution of the CLSP, a fundamental production planning problem in various industry settings. Our ML approach could be beneficial in reducing the solution time for many practical problems that are solved repeatedly with different parameters. We utilize bidirectional LSTMs to process information in both time directions. The metrics, defined as time factor improvement, infeasibility, and optimality gap, are presented to assess the quality of the predictions. The results for the CLSP instances with the same characteristics show that a time factor improvement of more than an order of magnitude can be achieved without much loss in feasibility or the optimality gap if the level of predictions used to solve the problem is well adjusted. Also, we tested if the trained LSTM models could generalize to instances with different data distributions or longer planning horizons. The results show that one should be careful in selecting the prediction level for predicting instances with different data distributions. The LSTM models trained on shorter planning horizons achieve great success in predicting instances with longer planning horizons with any prediction level and reduce the solution time up to six orders of magnitude with a small optimality gap. Specifically, we observe the highest computational benefit from our LSTM-Opt approach when predicting the hardest set of instances. Also, LSTM-Opt framework outperforms classical ML algorithms in terms of the quality of the solution and exact approaches with respect to the solution time improvement.

Our LSTM-Opt framework can be especially useful for reducing the solution time of dynamic combinatorial optimization problems that are solved in a repetitive setting. In this paper, we have used the CLSP as a specific case to show that deep learning approaches have great potential for learning optimal solutions to MIP problems. Future research could further investigate the generalizability of our approach to instances with a larger planning horizon and different distributions in more detail. Another possible research direction is to develop methods to eliminate infeasible predictions. Additionally, the developed LSTM-Opt framework can be extended to solve more complex versions of the CLSP, such as the multi-item or multi-level CLSP, as well as other sequential decision-making problems.

\section*{Acknowledgment}
We gratefully acknowledge the support of the National Science Foundation CAREER Award co-funded by the CBET/ENG Environmental Sustainability program and the Division of Mathematical Sciences in MPS/NSF under Grant No. CBET-1554018.


\renewcommand\refname{References} 
\bibliographystyle{apalike}
\bibliography{LotsizingLSTMReferences} 

\end{document}